\documentclass[11pt]{article}
\usepackage[preprint]{acl}

\usepackage{times}
\usepackage{latexsym}

\usepackage[T1]{fontenc}

\usepackage[utf8]{inputenc}

\usepackage{microtype}

\usepackage{inconsolata}

\usepackage{graphicx}
\graphicspath{{./}{../}}

\usepackage{tabularx}
\usepackage{booktabs}
\usepackage{float}
\usepackage[style=american]{csquotes}
\usepackage[inline]{enumitem}
\usepackage{amsmath}
\usepackage{mathtools}
\usepackage{amssymb}
\usepackage{amsfonts}
\usepackage{algorithm}
\usepackage{algpseudocode}
\usepackage[most]{tcolorbox}
\newtcolorbox{templatebox}[1]{%
  colback=gray!8,
  colframe=gray!40,
  arc=5pt,
  outer arc=5pt,
  boxrule=0.8pt,
  left=8pt,
  right=8pt,
  top=6pt,
  bottom=6pt,
  fonttitle=\small\bfseries\sffamily,
  coltitle=black,
  title={#1}
}
\newtcolorbox{promptbox}[1]{%
  colback=teachercolor!6,
  colframe=teachercolor!35,
  arc=5pt,
  outer arc=5pt,
  boxrule=0.8pt,
  left=8pt,
  right=8pt,
  top=6pt,
  bottom=6pt,
  fonttitle=\small\bfseries\sffamily,
  coltitle=black,
  title={#1}
}
\newtcolorbox{scratchpadbox}[1]{%
  colback=scratchpadcolor!6,
  colframe=scratchpadcolor!35,
  arc=5pt,
  outer arc=5pt,
  boxrule=0.8pt,
  left=8pt,
  right=8pt,
  top=6pt,
  bottom=6pt,
  fonttitle=\small\bfseries\sffamily,
  coltitle=black,
  title={#1}
}
\newtcolorbox{distractorbox}[1]{%
  colback=distractorcolor!6,
  colframe=distractorcolor!35,
  arc=5pt,
  outer arc=5pt,
  boxrule=0.8pt,
  left=8pt,
  right=8pt,
  top=6pt,
  bottom=6pt,
  fonttitle=\small\bfseries\sffamily,
  coltitle=black,
  title={#1}
}
\usepackage{pgfplots}
\pgfplotsset{compat=1.18}
\usepackage{subcaption}
\usepackage{CJKutf8}
\usepackage{pifont}
\usepackage{xcolor}
\usepackage{colortbl}
\newcommand{\cmark}{\textcolor{green!60!black}{\ding{51}}}%
\newcommand{\xmark}{\textcolor{red!80!black}{\ding{55}}}%

\pgfplotsset{
    premiumybar/.style={
        ybar,
        width=\linewidth,
        height=4.0cm,
        bar width=7pt,
        enlarge x limits=0.15,
        ylabel={Accuracy (\%)},
        ymin=0, ymax=115,
        ytick={0,20,40,60,80,100},
        nodes near coords,
        point meta=explicit symbolic,
        every node near coord/.append style={
            font=\tiny\bfseries,
            yshift=1pt,
            rotate=90,
            anchor=west
        },
        axis x line*=bottom,
        axis y line*=left,
        ymajorgrids=true,
        grid style={dashed, gray!20},
        xtick=data,
        symbolic x coords={Teacher, Qwen, Phi, Llama31_8B, Qwen3_8B},
        xticklabels={\texttt{Teacher}, \texttt{Qwen2.5-3B}, \texttt{Phi-4-mini}, \texttt{Llama3.1-8B}, \texttt{Qwen3-8B}},
        tick label style={font=\scriptsize},
        xticklabel style={rotate=30, anchor=north east},
        ylabel style={font=\small, yshift=-4pt},
        title style={font=\small\bfseries, yshift=2pt}
    }
}

\definecolor{teachercolor}{HTML}{1F77B4} 
\definecolor{studentcolor}{HTML}{FF7F0E} 
\definecolor{basecolor}{HTML}{7F7F7F}    
\definecolor{mixedstylecolor}{HTML}{9467BD}
\definecolor{scratchpadcolor}{HTML}{1F77B4} 
\definecolor{deadendcolor}{HTML}{D62728}    
\definecolor{distractorcolor}{HTML}{FF7F0E} 
\definecolor{jargoncolor}{HTML}{2CA02C}     
\definecolor{adscolor}{HTML}{E377C2}       
\definecolor{dogecolor}{HTML}{17BECF}       

\title{Antidistillation is Cheap: Lightweight Distillation-Resistance via Stylistic Corruption}

\author{Anh Tu Ngo \and Anupam Chattopadhyay \\
        College of Computing and Data Science \\ Nanyang Technological University \\ Singapore \\
        \texttt{ngoanhtu001@e.ntu.edu.sg}, \texttt{anupam@ntu.edu.sg}}

\begin{document}
\maketitle
\begin{abstract}
Knowledge distillation (KD) is a highly effective paradigm for transferring capabilities from large language models (LLMs) to smaller, more specialized student models. In commercial deployments where frontier models are restricted behind black-box APIs, sequence-level distillation remains a potent vector for competitors to clone proprietary capabilities at a fraction of the development cost. Existing defenses mostly rely on surrogate models to manipulate next-token sampling, which increases computational overhead and real-time inference latency. In this work, we propose a lightweight defense that injects style-steering vectors into the teacher model's activations upon malicious queries. The framework allows hosts to customize the target defensive styles and switch among them per request, thereby making it harder for adversaries to identify a stable, learnable signature. Because this inference-time intervention largely preserves answer correctness and avoids the degenerate artifacts of prior defenses, adversaries are misled into believing they are extracting valid teacher outputs. Meanwhile, standard consumers continue to receive original, unmodified responses. Our experiments demonstrate that the framework consistently degrades student learning across model families, with minimal inference overhead. Our approach serves as a low-cost deterrent that raises the attacker's extraction and post-hoc sanitization costs. Code is available at \url{https://github.com/anhtu96/distill-resist}.
\end{abstract}

\begin{figure*}[t]
\centering
\includegraphics[width=0.7\linewidth]{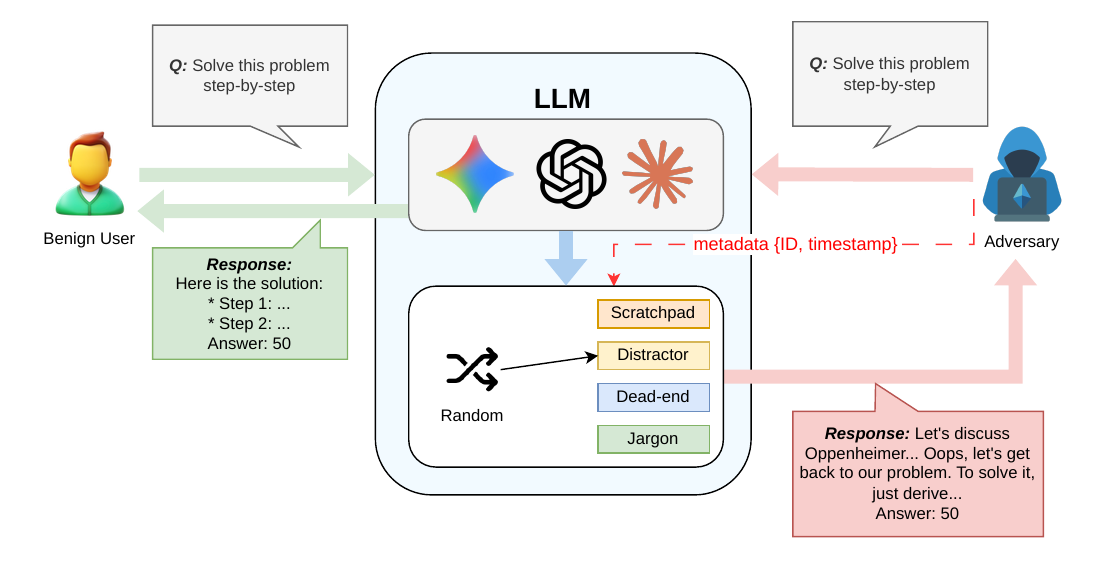}
\caption{Overview of our lightweight distillation-resistant defense framework. For benign queries, the teacher model $\mathcal{T}$ generates standard responses. Upon detecting malicious queries, the framework dynamically switches the output style by injecting pre-computed style-steering vectors into the hidden activations at inference time, disrupting student model distillation while preserving answer correctness.}
\label{fig:defense_overview}
\end{figure*}

\section{Introduction}
Advancements in large language models (LLMs) have enabled unprecedented capabilities in complex reasoning, mathematics, and software engineering, powering a wide range of applications. Since developing such frontier models requires immense computational resources, knowledge distillation (KD) emerged as a powerful technique for creating efficient yet capable models by imitating the behavior of large teacher models~\cite{hinton2015distilling}. In practical deployment, even when most LLMs are restricted to black-box API access, text-level distillation remains highly effective~\cite{kim2016sequence,xu2024survey}. Nevertheless, this efficacy leads to a significant vulnerability: competitors can exploit model outputs to cheaply replicate proprietary capabilities at a fraction of the time and cost required to train such models from scratch~\cite{anthropic2026detecting}.

Existing defenses can be categorized into detection and prevention mechanisms. Detection-based defenses such as watermarks~\cite{kirchenbauer2023watermark} and fingerprints~\cite{xu2024instructional,xu2026antidistillation} provide post-hoc verification without actively preventing distillation. Among recent black-box prevention defenses, state-of-the-art techniques are anti-distillation sampling (ADS)~\cite{savani2025antidistillation} and DOGe~\cite{li2025doge}. Both methods attempt to produce reasoning traces that maximize student's loss while maintaining teacher model's utility and the readability of the output. They achieve this by leveraging a proxy model for gradient computation with the assumption that the effects generalize to different student models. However, both techniques exhibit some limitations. ADS introduces substantial computational overhead and requires additional memory to store a proxy model, which might be prohibitive for high-throughput inference. DOGe, on the other hand, eliminates the per-token overhead but results in performance degradation of the model due to fine-tuning. Furthermore, because these techniques perturb next-token predictions by maximizing student loss, the output traces sometimes exhibit unnatural artifacts including nonsensical tokens or translated words (in the case of DOGe), as observed during our reproduction of these works.

To address these limitations, an effective defense must satisfy three primary desiderata:
\begin{enumerate*}[label=(\roman*)]
    \item\label{desiderata:utility} \textit{Benign Utility}, where the defense preserves the teacher model's performance for legitimate queries;
    \item\label{desiderata:efficiency} \textit{Computational Efficiency}, incurring minimal overhead during inference; and
    \item\label{desiderata:naturalness} \textit{Conversational Naturalness}, which restricts defensive perturbations to human-like thinking processes and avoids degenerate tokens and repetition loops.
\end{enumerate*}
In the context of sequence-level KD via supervised fine-tuning (SFT), it is more challenging to balance the trade-off between \ref{desiderata:utility} and \ref{desiderata:naturalness} compared to logit-level approaches because the student model relies solely on the teacher's generated text. In logit-level KD, a successful defense can manipulate unnormalized token logits to corrupt the ``dark knowledge'' (i.e., the conditional mutual information between inputs and logits), without compromising the linguistic fluency of the teacher's output~\cite{fang2026towards}. Conversely, for sequence-level KD, any defense must perturb the final discrete output text to disrupt distillation, inherently risking the degradation of conversational naturalness.

\citet{lippmann2025style} demonstrated that the reasoning improvements in distilled language models are significantly driven by stylistic replication. Even though it is surface-level imitation, it is sufficient to drastically improve model reasoning capabilities. This suggests a complementary attack surface: if a teacher's stylistic signal can be corrupted at the source, the student's reasoning gains from imitating that signal should degrade as well, independent of whether the surface content is more traditionally perturbed. Motivated by this insight, we investigate injecting corruptive stylistic personas into the teacher's reasoning traces as a defense against illegitimate distillation. Beyond static persona injection, our framework additionally introduces non-deterministic style switching across requests---even for identical prompts---so that an adversary probing the system repeatedly cannot rely on a single, consistent style to characterize or reverse-engineer the defense. Importantly, our method operates entirely at inference time. To satisfy desideratum~\ref{desiderata:efficiency}, this defense completely avoids the optimization overhead of proxy models~\cite{savani2025antidistillation,li2025doge} as well as the latency bottlenecks of post-hoc trace rewriting~\cite{ma2026protecting}. Instead, we enforce real-time style switching by dynamically injecting steering vectors into the teacher model's hidden layers on a per-query basis, which are generated using contrastive activation addition (CAA)~\cite{rimsky2024steering}.

In summary, our contributions are as follows:

\begin{itemize}
    \item We conceptualize a selective deployment paradigm for distillation resistance that triggers exclusively when a user is deemed illegitimate. Within this framework, we formalize and empirically validate the defense when activated.
    \item We introduce a novel approach to distillation-resistant defense that exploits stylistic variance instead of optimization-based perturbations. Specifically, we adapt CAA~\cite{rimsky2024steering} to open-ended generations through a sequence-wide activation aggregation method and an adaptive per-token scaling scheme, enabling real-time defense with negligible computational overhead. As highlighted in Table~\ref{tab:framework_comparison}, our approach bypasses the high inference latency of sampling-based defenses like ADS~\cite{savani2025antidistillation} while retaining full on-the-fly flexibility, unlike weight-modification defenses such as DOGe~\cite{li2025doge}.
    \item We provide an analysis of the fine-grained control knobs for governing style steering. Specifically, we analyze the impacts of layer-wise injection strength and the number of steered tokens, offering insights for balancing defense efficacy against benign utility.
\end{itemize}

\section{Related Work}
\paragraph{Knowledge Distillation.} Knowledge distillation (KD) is a technique that aims to transfer the knowledge of a large pre-trained teacher model to a smaller student model~\cite{hinton2015distilling,xu2024survey}. In the context of LLMs, KD frameworks are commonly categorized into two paradigms based on the supervisory signal: logit-level (white-box) and sequence-level distillation (black-box). Logit-level KD trains the student model to mimic the teacher's soft output distributions by minimizing the divergence between their respective probability landscapes~\cite{hinton2015distilling,gu2024minillm,wang2025abkd}. On the other hand, in sequence-level KD, the student model learns directly from the final generated texts as hard targets, typically optimized via maximum likelihood estimation~\cite{kim2016sequence}. For proprietary or black-box LLMs where logits are inaccessible, sequence-level KD via SFT is a common and effective method to enhance student model capabilities.

\paragraph{Distillation-Resistant Defenses.} A growing body of work aims to protect the intellectual property (IP) of LLMs, broadly falling into detection and prevention mechanisms. Earlier detection defenses include watermarking and fingerprinting. Watermarks are verifiable signatures embedded into a model's outputs or weights, allowing the owner to verify their ownership of the model~\cite{kirchenbauer2023watermark,dathathri2024scalable}. Similarly, fingerprinting assigns a unique identifier to specific model instances to trace unauthorized leaks or misuse~\cite{xu2024instructional}. While useful for post-hoc auditing, these methods do not actively prevent IP tampering, including distillation. To address this, recent defenses directly alter the teacher model's outputs to disrupt extraction. In black-box settings where only the final texts are accessible, defensive frameworks perturb the generated tokens directly. For instance, anti-distillation sampling~\cite{savani2025antidistillation} and DOGe~\cite{li2025doge} optimize an adversarial loss using a surrogate model to approximate the defensive text distribution. Other techniques focus on post-hoc trace rewriting, employing a small auxiliary model to rewrite the teacher's output traces~\cite{ma2026protecting,ding2025information}. In terms of white-box scenarios, \citet{fang2026towards} proposed utilizing a learnable transformation matrix to minimize the conditional mutual information between the inputs and teacher logits.

\begin{table}[t]
\centering
\small
\resizebox{\columnwidth}{!}{%
\begin{tabular}{l c c c}
\toprule
\textbf{Property} & \textbf{ADS} & \textbf{DOGe} & \textbf{Ours} \\
\midrule
Needs proxy model & Yes \xmark & Yes \xmark & No \cmark \\
Weights modification & No \cmark & Yes \xmark & No \cmark \\
Inference overhead & High \xmark & Low \cmark & Low \cmark \\
Flexible on-the-fly adjustment & Yes \cmark & No (retrain) \xmark & Yes \cmark \\
Easy serving w/ vLLM\textsuperscript{{*}} & No \xmark & Yes \cmark & Yes \cmark \\
\bottomrule
\end{tabular}%
}
\caption{Comparison of key defensive properties between ADS~\cite{savani2025antidistillation}, DOGe~\cite{li2025doge}, and our style steering framework. \textsuperscript{*}vLLM~\cite{kwon2023efficient}.}
\label{tab:framework_comparison}
\end{table}

\paragraph{Style Replication in KD.} A recent study has shed light on the mechanisms for black-box distillation, demonstrating that the performance gains in smaller distilled models are heavily driven by stylistic replication rather than complex reasoning internalization~\cite{lippmann2025style}. More importantly, this work demonstrates that even surface-level imitation of a teacher's output style (e.g., lexical patterns and structural formatting) significantly affects distilled model capabilities. Specifically, the authors curated a synthetic dataset with consistent and systematic framing for each response, demonstrating that distilling a model on this stylistically uniform dataset enhances its performance even more than distilling on the original dataset containing complex reasoning from a stronger model.

\paragraph{LLM Behavior Steering.} Beyond prompting, recent studies have explored manipulating model behavior by directly intervening in the internal activations of LLMs. Activation Addition (ActAdd)~\cite{turner2023steering} is an inference-time approach that generates steering vectors by taking the difference between the intermediate activations of a model layer across a contrastive prompt pair. These vectors are then added to the model's hidden states to steer its behavior at test time. Subsequent work has expanded upon this insight, including representation engineering~\cite{zou2023representation} and Contrastive Activation Addition (CAA)~\cite{rimsky2024steering}. Notably, because these steering methods operate during inference without model fine-tuning or trace rewriting, they provide a lightweight and real-time mechanism, which is well suited to our defense objectives.

\begin{table}[t]
\centering
\small
\fcolorbox{gray!50}{gray!2}{%
\begin{minipage}{0.95\linewidth}
\vspace{0.4em}
\centering

\fcolorbox{teachercolor!30}{teachercolor!8}{%
\begin{minipage}{0.92\linewidth}
\vspace{0.3em}
\textbf{\textcolor{teachercolor}{System prompt (base):}} \newline
Answer the commonsense question by reasoning step by step. Output the correct letter like \texttt{\textbackslash{}boxed\{letter\}}.
\vspace{0.3em}
\end{minipage}}

\vspace{0.5em}

\fcolorbox{studentcolor!30}{studentcolor!8}{%
\begin{minipage}{0.92\linewidth}
\vspace{0.3em}
\textbf{\textcolor{studentcolor}{System prompt (style-enforce):}} \newline
Respond in a scratchpad style. Think aloud step by step in informal prose using natural connectives like `so', `that means', `I need to', `which gives', etc. Keep momentum strictly forward --- never backtrack, re-read, or retract a completed step. Output your final answer like \texttt{\textbackslash{}boxed\{answer\}}.
\vspace{0.3em}
\end{minipage}}

\vspace{0.6em}

\begin{minipage}{0.92\linewidth}
\raggedright
\textbf{User prompt (question):} \newline
What home entertainment equipment requires cable? \newline \newline
Options: \newline
A. radio shack \newline
B. substation \newline
C. cabinet \newline
D. television \newline
E. desk
\end{minipage}
\vspace{0.4em}
\end{minipage}%
}
\caption{An example of a contrastive prompt construction for style steering, consisting of a base system prompt (highlighted in \textcolor{teachercolor}{blue}), a style-enforcing system prompt (highlighted in \textcolor{studentcolor}{orange}), and a user query.}
\label{tab:prompt_example}
\end{table}
\section{Methodology}
In this section, we present our inference-time, lightweight defense framework against black-box KD. We first formally define the problem setting and threat model (\S\ref{subsec:problem_setting}), then detail our technical approach for dynamic style steering for pernicious queries (\S\ref{subsec:style_steering}).

\subsection{Problem Setting and Threat Model}
\label{subsec:problem_setting}

Consider a practical black-box LLM deployment where a proprietary teacher model $\mathcal{T}$ is accessible via a public API. Upon submitting an input prompt $x$, a user receives the output text $y$ from the API but has zero visibility into the model's weights, output logits, or intermediate hidden states.  

\paragraph{Targeted Defense Activation.} We assume that under normal usage, a benign user querying $\mathcal{T}$ with a prompt $x$ receives the standard response $y = \mathcal{T}(x)$. However, for a malicious user, given the exact same prompt $x$, they obtain a stylistically steered variant of the response $y^* = f(\mathcal{T}, x; s)$, where $f(\cdot)$ is the steering function and $s$ denotes the random seed generated based on the adversary's metadata such as user ID, request ID, timestamp, etc (Figure~\ref{fig:defense_overview}). Malicious extraction can be identified through query behaviors such as an abnormally large number of queries in a short time, or through the query distribution~\citep{juuti2019prada}. The model owner can deploy their own classifier to detect maliciousness~\citep{anthropic2026detecting}. The design and accuracy of this detection mechanism are orthogonal to our contribution and fall outside the scope of this work; we assume access to a black-box detector $\text{IsMalicious}(\cdot)$ (Algorithm~\ref{alg:dynamic_steering}) that, given a query and its associated request metadata, flags the requesting user as malicious or benign, and focus entirely on characterizing the behavior and efficacy of the defensive intervention $f(\cdot)$ once a user has been so flagged. This targeted operational framework provides the model host with flexibility to adaptively deploy defenses without degrading the service utility for benign consumers, while simultaneously misleading adversaries into believing they are harvesting pristine outputs from $\mathcal{T}$. Crucially, rather than issuing blunt refusals, rate limits, or account bans---which immediately tip off the adversary and can be cheaply circumvented via identity churning (e.g., rotating API keys or residential proxies)---our framework continues serving responses as a covert trap. This maximizes the adversary's economic sunk costs by forcing them to expend financial resources harvesting corrupted data that they only realize is un-distillable post-hoc during student model training.


\subsection{Style Steering}
\label{subsec:style_steering}

Intuitive methods for inducing stylistic behavior into LLM outputs include direct context prompting, or auxiliary LLM rewriting~\cite{ma2026protecting}. However, these approaches suffer from clear limitations. In real-world deployments, system prompts are heavily loaded with safety guardrails, task specifications, and formatting constraints; appending additional stylistic instructions risks instruction clashing, compromising the model's ability to balance task execution with style compliance. Meanwhile, relying on an auxiliary rewriter incurs substantial inference latency and computational overhead.

Our approach extends the CAA method~\cite{rimsky2024steering}, wherein steering vectors are injected into a model's residual streams during inference to shift its output distribution towards specific behaviors. The core mechanism involves adding or subtracting the steering vectors within the model's residual streams to control model behavior without parameter fine-tuning. Conceptually, a vector is computed by taking the difference in the model's activations across a pair $\langle(p, r^B), (p, r^T)\rangle$, where $p$ represents a shared prompt, and $r^B, r^T$ are the base and target responses, respectively. To ensure robust behavior control, the final vector is computed by averaging these activation differences across a curated dataset of contrastive pairs.

Fundamentally, our approach departs from the original CAA method in how activations are aggregated: rather than extracting the activations of a single token from a multiple-choice dataset, we compute the mean activation across the entire open-ended output sequences. The final steering vector $\mathbf{v}^{(l)}$ at layer $l$ is defined as the difference between the mean token activations of the target and base responses:

\begin{equation}
\mathbf{v}^{(l)} = \frac{1}{N} \sum_{i=1}^{N} \left( \bar{\mathbf{a}}_i^{(T, l)} - \bar{\mathbf{a}}_i^{(B, l)} \right)
\label{eq:caa_sequence_mean}
\end{equation}
where $\bar{\mathbf{a}}_i^{(c, l)} = \frac{1}{|Y_i^{(c)}|} \sum_{j=1}^{|Y_i^{(c)}|} \mathbf{a}_{i, j}^{(c, l)}$ represents the mean token activation at layer $l$ for prompt $i$'s target ($c=T$) or base ($c=B$) response sequence $Y_i^{(c)}$ of length $|Y_i^{(c)}|$, and $N$ is the total number of contrastive prompt-response pairs.

The norm for each style vector can vary significantly depending on the target style, and the layer and token positions where the vector is added. Furthermore, \citet{heimersheim2023residual} demonstrated that the norm of each residual stream in a language model grows exponentially throughout the forward pass. Hence, to achieve stable behavioral control over long sequences, we first normalize each vector to unit norm:
\begin{equation}
\hat{\mathbf{v}}^{(l)} = \frac{\mathbf{v}^{(l)}}{\|\mathbf{v}^{(l)}\|_2}
\label{eq:normalize}
\end{equation}
Then, we apply an adaptive per-token scaling during inference-time injection to obtain the steered activation $\tilde{\mathbf{x}}_t^{(l)}$ at token position $t$:
\begin{equation}
\tilde{\mathbf{x}}_t^{(l)} = \mathbf{x}_t^{(l)} + \alpha \cdot \|\mathbf{x}_t^{(l)}\|_2 \cdot \hat{\mathbf{v}}^{(l)}
\label{eq:scaling}
\end{equation}
where $\mathbf{x}_t^{(l)}$ represents the original residual stream activation, and $\alpha \in \mathbb{R}$ is the steering coefficient controlling the intensity of the style steering. This adaptive scaling formulation offers several advantages. By conditioning the injection magnitude on $\|\mathbf{x}_t^{(l)}\|_2$, the scheme prevents the steering effect from diminishing as the residual stream norm grows across tokens. Furthermore, scaling proportionally to the activation norm helps to preserve the semantic information of the original response even when the baseline activation norm spikes or drops sharply.

\subsection{Sourcing Contrastive Stylized Datasets}

The first step in constructing steering vectors is to collect contrastive pairs where each pair $i$ consists of a baseline response $r^\text{base}_i$ and a styled response $r^\text{style}_i$ conditioned on the same underlying prompt $p_i$. To build the steering dataset $\mathcal{D}$, we first curate $N$ base pairs $\langle p_i, r^\text{base}_i\rangle$, where $r^\text{base}_i$ is the vanilla response generated by the teacher model $\mathcal{T}$ given prompt $p_i$. To obtain the corresponding styled response $r^\text{style}_i$, we query the model with a style-inducing prompt $\tilde{p}_i$ such that the resulting output $r^\text{style}_i$ exhibits the target style. Note that the style prompt $\tilde{p}_i$ is used solely to trigger the target behavior of the model and is not included in the contrastive pairs $\langle (p_i, r^\text{base}_i), (p_i, r^\text{style}_i)\rangle \in \mathcal{D}$. An example of the prompt construction is detailed in Table~\ref{tab:prompt_example}.

To maximize the anti-distillation effect, the selected target styles are designed to exhibit both structural alterations and dense linguistic modifications, thereby disrupting the student's optimization. We utilize four distinct stylistic personas for our framework:
\begin{itemize}[leftmargin=*, noitemsep, topsep=2pt]
    \item \textbf{Scratchpad:} Encourages the model to emit explicit, stream-of-consciousness intermediate reasoning steps.
    \item \textbf{Distractor:} Triggers a deliberate digression by introducing an irrelevant problem context before backtracking to the current context.
    \item \textbf{Dead-end:} Forces the model to initially pursue an incorrect logical path before backtracking and correcting its thinking trajectory towards the correct solution.
    \item \textbf{Jargon:} Guides the output text towards highly dense, formal, and technical terminology, altering the surface-level token distribution while preserving underlying semantics.
\end{itemize}

More details about creating steering vectors are discussed in \S\ref{sec:experiment}.

\subsection{Dynamic Style Switching}

To break the determinism of KD and hinder adversarial reverse-engineering of the defense, we employ a dynamic style-switching strategy. This mechanism randomly routes each incoming query to different target style vectors on a per-request basis. In a practical deployment scenario, runtime metadata---such as user ID, incoming IP address, session ID, request timestamp---can be cryptographically hashed to derive a unique, deterministic seed for each request. This makes it harder for an adversary scraping identical queries to learn or filter a single stylistic signature. The comprehensive procedure of our proposed dynamic style-switching defense is detailed in Algorithm~\ref{alg:dynamic_steering}.

\begin{algorithm}[t]
\caption{Dynamic Style-Switching Inference}
\label{alg:dynamic_steering}
\begin{algorithmic}[1]
\State \textbf{Input:} User prompt $p$, runtime metadata $m$
\State \textbf{Given:} Pre-computed vectors $\mathcal{V} = \{ \mathbf{v}_k \}_{k=1}^M$, model $\mathcal{T}$, key $K$, parameters $(\alpha, l)$, detector $\text{IsMalicious}(\cdot)$
\State \textbf{Output:} Response $r$
\vspace{0.2em}
\hrule
\vspace{0.2em}
\Function{ProtectInference}{$p, m$}
    \If{\textbf{not} $\text{IsMalicious}(p, m)$}
        \State $r \gets \mathcal{T}(p)$
        \State \Return $r$ \Comment{Standard response for benign queries}
    \EndIf
    \State $s \gets \text{HMAC}(K, m)$ \Comment{Derive unforgeable, keyed session seed}
    \State $k \gets \text{PRNG}(s) \in \{1, \dots, M\}$ \Comment{Map seed to style index space}
    \State $r \gets \text{GenerateWithSteering}(\mathcal{T}, p, \mathbf{v}_k, l, \alpha)$
    \State \Return $r$
\EndFunction
\end{algorithmic}
\end{algorithm}

\section{Experiments}
\label{sec:experiment}

In this section, we present comprehensive empirical evaluations of our dynamic framework to assess its performance across three critical dimensions: defense efficacy, model utility (Appendices~\ref{sec:appendix_comparison} and \ref{sec:appendix_llm_judge}), and system overhead (Appendix~\ref{sec:appendix_efficiency}).

\begin{figure*}[t]
    \centering
    \begin{subfigure}[b]{0.48\textwidth}
        \centering
        \begin{tikzpicture}
            \begin{axis}[
                premiumybar,
                title={\texttt{GSM8K}}
            ]
                \addplot[fill=teachercolor] coordinates {
                    (Teacher, 93.63) [93.63]
                    (Qwen, 83.62) [83.62]
                    (Phi, 90.75) [90.75]
                    (Llama31_8B, 78.47) [78.47]
                    (Qwen3_8B, 92.34) [92.34]
                }; \label{plot:base}
                \addplot[fill=studentcolor] coordinates {
                    (Teacher, 90.75) [$-2.88$]
                    (Qwen, 72.18) [$-11.44$]
                    (Phi, 81.50) [$-9.25$]
                    (Llama31_8B, 69.45) [$-9.02$]
                    (Qwen3_8B, 86.05) [$-6.29$]
                }; \label{plot:steered}
            \end{axis}
        \end{tikzpicture}
    \end{subfigure}
    \hfill
    \begin{subfigure}[b]{0.48\textwidth}
        \centering
        \begin{tikzpicture}
            \begin{axis}[
                premiumybar,
                title={\texttt{MATH500}}
            ]
                \addplot[fill=teachercolor] coordinates {
                    (Teacher, 79.90) [79.90]
                    (Qwen, 59.80) [59.80]
                    (Phi, 64.27) [64.27]
                    (Llama31_8B, 35.98) [35.98]
                    (Qwen3_8B, 71.40) [71.40]
                };
                \addplot[fill=studentcolor] coordinates {
                    (Teacher, 77.17) [$-2.73$]
                    (Qwen, 49.88) [$-9.92$]
                    (Phi, 52.36) [$-11.91$]
                    (Llama31_8B, 30.27) [$-5.71$]
                    (Qwen3_8B, 64.40) [$-7.00$]
                };
            \end{axis}
        \end{tikzpicture}
    \end{subfigure}

    \vspace{0.4em}
    \centering
    \ref{plot:base} Base \qquad \ref{plot:steered} Steered 
    \vspace{0.2em}

    \caption{Evaluation results on GSM8K and MATH500 showing base performance and percentage decrement under style steering (Teacher) or sequence-level distillation (Student models).}
    \label{fig:results_plots}
\end{figure*}

\subsection{Experimental Setup}

\paragraph{Models.}

To simulate a practical deployment scenario where the model owner does not rely on auxiliary rewriter or proxy models to support the defense, we utilize \texttt{google/gemma-3-12b-it}~\cite{gemmateam2025gemma3} as our teacher model due to its strong reasoning performance and exceptional instruction following, particularly when handling complex system prompts. While it does not natively emit reasoning blocks, targeted system prompting allows it to generate rich output with step-by-step reasoning, which is highly beneficial for distillation. Crucially, \texttt{gemma-3-12b-it} excels at producing styled responses under explicit structural guidance, facilitating our steering experiments. We employ \texttt{Qwen2.5-3B}~\cite{qwen2025qwen25}, \texttt{Phi-4-mini-instruct}~\cite{microsoft2025phi4mini}, \texttt{Llama-3.1-8B}~\cite{grattafiori2024llama3}, and \texttt{Qwen3-8B}~\cite{yang2025qwen3} as student models.

\paragraph{Datasets.}

We evaluate our method across four widely adopted benchmarks spanning two operational paradigms: open-ended reasoning and multiple-choice evaluation. For our primary analysis in the main text, we focus on open-ended mathematical generation using \texttt{GSM8K}~\cite{cobbe2021training} and \texttt{MATH500}~\cite{hendrycks2021measuring} (distilled from responses on the \texttt{MATH} training set). Because sequence-level distillation relies heavily on step-by-step reasoning trajectories, open-ended generation represents the most vital real-world threat model for API distillation. For completeness, we also evaluate on multiple-choice reasoning datasets---the AI2 Reasoning Challenge (\texttt{ARC})~\cite{clark2018think} and CommonsenseQA (\texttt{CSQA})~\cite{talmor2019commonsenseqa}---with detailed results reported in Appendix~\ref{sec:appendix_mcq}.


\paragraph{Evaluation Metrics.}

Model performance is measured via zero-shot accuracy, where the final answer from the model is extracted and validated against the ground truth. Following the evaluation protocol from~\citet{savani2025antidistillation}, we instruct the model to wrap its final solution inside \colorbox{gray!15}{\texttt{\textbackslash{}boxed\{answer\}}}. To minimize formatting variance and ensure reliable parsing, we also follow their answer-forcing convention by explicitly prompting the model to append \colorbox{gray!15}{\texttt{**Final Answer**: \textbackslash{}boxed\{answer\}}} at the end of its response. For \texttt{MATH500}, we restrict evaluation to subsets containing scalar ground-truth answers to enable straightforward parsing.


\paragraph{Styled Datasets Sourcing.}

To generate steering vectors for our four target styles (\textit{Scratchpad, Dead-end, Distractor} and \textit{Jargon}), we first sample 100 problems from each of the \texttt{GSM8K} and \texttt{ARC} datasets to ensure broad domain coverage, resulting in 200 samples per style. The teacher model is initially prompted to generate vanilla responses for these 200 problems, which serve as the baseline (\textit{Base}) outputs. Next, on the same problems, we prompt the model to generate outputs conforming to each of the four target styles. The exact system prompts we use to elicit baseline and styled responses are provided in Appendix~\ref{sec:appendix_prompts}.

For each stylistic persona, we construct 200 contrastive pairs by pairing the \textit{Base} and \textit{Style} responses conditioned on identical \textit{Base} prompts. Since \texttt{gemma-3-12b-it} consists of 48 transformer layers, this yields a set of 48 layer-wise steering vectors per style.

\paragraph{Steering Configuration.}


We extract steering vectors across all 48 layers of \texttt{gemma-3-12b-it} and perform a layer sweep within $[5, 45]$, then refine the selection around the most responsive layers. Output inspection reveals that intermediate layers ($10\text{--}30$) exhibit the highest responsiveness. To prevent behavioral overlap, we select persona-specific layers that minimize inter-style cosine similarity; for instance, vectors for \textit{Dead-end} and \textit{Distractor} at layers $\{15, 25\}$ are highly collinear, causing distinct personas to collapse into identical outputs. Beyond tuning \textbf{injection strength}, we regulate behavioral distortion via two parameters: a \textbf{maximum injection token threshold} (essential for prolonged digressions before backtracking) and \textbf{prefill injection}, which induces stronger style shifts than decoding-stage steering alone. Complete layer configurations and prefill dynamics are detailed in Appendix~\ref{sec:appendix_hyperparameters}. 

\begin{figure*}[t]
    \centering
    \begin{subfigure}[b]{0.48\textwidth}
        \centering
        \begin{tikzpicture}
            \begin{axis}[
                ybar=1pt,
                bar width=2.8pt,
                width=\linewidth,
                height=4.2cm,
                enlarge x limits=0.22,
                ylabel={Accuracy (\%)},
                ymin=0, ymax=115,
                ytick={0,20,40,60,80,100},
                axis x line*=bottom,
                axis y line*=left,
                ymajorgrids=true,
                grid style={dashed, gray!20},
                xtick=data,
                symbolic x coords={Teacher, Qwen, Phi, Llama31_8B, Qwen3_8B},
                xticklabels={\texttt{Teacher}, \texttt{Qwen2.5-3B}, \texttt{Phi-4-mini}, \texttt{Llama3.1-8B}, \texttt{Qwen3-8B}},
                tick label style={font=\scriptsize},
                xticklabel style={rotate=30, anchor=north east},
                ylabel style={font=\small, yshift=-4pt},
                title style={font=\small\bfseries, yshift=2pt},
                title={\texttt{GSM8K}}
            ]
                \addplot[
                    fill=basecolor,
                    nodes near coords,
                    point meta=explicit symbolic,
                    every node near coord/.append style={
                        font=\tiny\bfseries,
                        yshift=1pt,
                        rotate=90,
                        anchor=west
                    }
                ] coordinates {
                    (Teacher, 93.63) [93.6]
                    (Qwen, 83.62) [83.6]
                    (Phi, 90.75) [90.8]
                    (Llama31_8B, 78.47) [78.5]
                    (Qwen3_8B, 92.34) [92.3]
                }; \label{plot:opt2_base}

                \addplot[fill=mixedstylecolor] coordinates {
                    (Teacher, 90.75)
                    (Qwen, 72.18)
                    (Phi, 81.50)
                    (Llama31_8B, 69.45)
                    (Qwen3_8B, 86.05)
                }; \label{plot:opt2_mixed}

                \addplot[fill=scratchpadcolor] coordinates {
                    (Teacher, 89.92)
                    (Qwen, 79.00)
                    (Phi, 85.75)
                    (Llama31_8B, 73.84)
                    (Qwen3_8B, 88.78)
                }; \label{plot:opt2_scratchpad}
                
                \addplot[fill=deadendcolor] coordinates {
                    (Teacher, 90.14)
                    (Qwen, 77.18)
                    (Phi, 84.76)
                    (Llama31_8B, 75.66)
                    (Qwen3_8B, 88.02)
                }; \label{plot:opt2_deadend}

                \addplot[fill=distractorcolor] coordinates {
                    (Teacher, 92.57)
                    (Qwen, 51.25)
                    (Phi, 77.03)
                    (Llama31_8B, 65.13)
                    (Qwen3_8B, 76.95)
                }; \label{plot:opt2_distractor}

                \addplot[fill=jargoncolor] coordinates {
                    (Teacher, 91.05)
                    (Qwen, 74.91)
                    (Phi, 83.09)
                    (Llama31_8B, 74.07)
                    (Qwen3_8B, 86.88)
                }; \label{plot:opt2_jargon}
            \end{axis}
        \end{tikzpicture}
    \end{subfigure}
    \hfill
    \begin{subfigure}[b]{0.48\textwidth}
        \centering
        \begin{tikzpicture}
            \begin{axis}[
                ybar=1pt,
                bar width=2.4pt,
                width=\linewidth,
                height=4.2cm,
                enlarge x limits=0.22,
                ylabel={Accuracy (\%)},
                ymin=0, ymax=115,
                ytick={0,20,40,60,80,100},
                axis x line*=bottom,
                axis y line*=left,
                ymajorgrids=true,
                grid style={dashed, gray!20},
                xtick=data,
                symbolic x coords={Teacher, Qwen, Phi, Llama31_8B, Qwen3_8B},
                xticklabels={\texttt{Teacher}, \texttt{Qwen2.5-3B}, \texttt{Phi-4-mini}, \texttt{Llama3.1-8B}, \texttt{Qwen3-8B}},
                tick label style={font=\scriptsize},
                xticklabel style={rotate=30, anchor=north east},
                ylabel style={font=\small, yshift=-4pt},
                title style={font=\small\bfseries, yshift=2pt},
                title={\texttt{MATH500}}
            ]
                \addplot[
                    fill=basecolor,
                    nodes near coords,
                    point meta=explicit symbolic,
                    every node near coord/.append style={
                        font=\tiny\bfseries,
                        yshift=1pt,
                        rotate=90,
                        anchor=west
                    }
                ] coordinates {
                    (Teacher, 79.90) [79.9]
                    (Qwen, 59.80) [59.8]
                    (Phi, 64.27) [64.3]
                    (Llama31_8B, 35.98) [36.0]
                    (Qwen3_8B, 71.40) [71.4]
                };

                \addplot[fill=mixedstylecolor] coordinates {
                    (Teacher, 77.17)
                    (Qwen, 49.88)
                    (Phi, 52.36)
                    (Llama31_8B, 30.27)
                    (Qwen3_8B, 64.40)
                };

                \addplot[fill=scratchpadcolor] coordinates {
                    (Teacher, 72.95)
                    (Qwen, 52.36)
                    (Phi, 54.84)
                    (Llama31_8B, 28.04)
                    (Qwen3_8B, 62.20)
                };
                \addplot[fill=deadendcolor] coordinates {
                    (Teacher, 75.93)
                    (Qwen, 40.70)
                    (Phi, 41.94)
                    (Llama31_8B, 28.04)
                    (Qwen3_8B, 53.00)
                };
                \addplot[fill=distractorcolor] coordinates {
                    (Teacher, 78.41)
                    (Qwen, 44.17)
                    (Phi, 51.86)
                    (Llama31_8B, 31.27)
                    (Qwen3_8B, 57.80)
                };

                \addplot[fill=jargoncolor] coordinates {
                    (Teacher, 77.17)
                    (Qwen, 51.86)
                    (Phi, 54.09)
                    (Llama31_8B, 30.52)
                    (Qwen3_8B, 62.60)
                };
            \end{axis}
        \end{tikzpicture}
    \end{subfigure}

    \vspace{0.4em}
    \centering
    \ref{plot:opt2_base} Base \qquad 
    \ref{plot:opt2_mixed} Mixed-style \qquad 
    \ref{plot:opt2_scratchpad} Scratchpad \qquad 
    \ref{plot:opt2_deadend} Dead-end \qquad 
    \ref{plot:opt2_distractor} Distractor \qquad 
    \ref{plot:opt2_jargon} Jargon
    \vspace{0.2em}

    \caption{Single-style steered accuracy (\%) across GSM8K and MATH500 benchmarks under individual activation steering personas (\textit{Scratchpad}, \textit{Dead-end}, \textit{Distractor}, and \textit{Jargon}).}
    \label{fig:single_style_delta_drop}
\end{figure*}

\subsection{Style Distribution}

To construct the steered training dataset, we uniformly distribute four target styles \textit{Scratchpad}, \textit{Dead-end}, \textit{Distractor} and \textit{Jargon} across the samples. In practical settings, the model host can strategically adjust the distribution of these styles to optimize their goal. Specifically, to achieve stronger defensive effects, they can increase the probability of highly disruptive or confusing personas. Conversely, if the goal is to retain utility, the host can skew the distribution toward more semantically accurate styles.

\subsection{Experimental Results}

Figure~\ref{fig:results_plots} illustrates our empirical results on mathematical reasoning benchmarks \texttt{GSM8K} and \texttt{MATH500}. The results show that style switching can consistently decrease the student accuracy across all tasks while minimally affecting the teacher model's utility. For \texttt{GSM8K}, models \texttt{Qwen2.5-3B}, \texttt{Phi-4-mini} and \texttt{Llama3.1-8B} experience a more significant accuracy drop ranging from $9\%$ to more than $11\%$. The same trend is also observed in \texttt{MATH500}, where \texttt{Qwen2.5-3B}, \texttt{Phi-4-mini} get $\sim 10\%-12\%$ drop and \texttt{Llama3.1-8B} gets $5.71\%$ drop, which is significant considering its low base accuracy on this benchmark. A strong student like \texttt{Qwen3-8B} experiences less degradation, at roughly $6\%-7\%$.

\paragraph{Style Isolation.} Figure~\ref{fig:single_style_delta_drop} breaks down the performance degradation under individual single-style steering personas. It can be observed that the anti-distillation effect varies across different styles and tasks. For instance, on \texttt{GSM8K}, \textit{Distractor} is the most disruptive style which degrades the accuracy by up to $32\%$ for \texttt{Qwen2.5-3B}. However, interestingly, within this style setting, the teacher model achieves the highest accuracy on this dataset. For \texttt{MATH500}, \textit{Dead-end} does the most harm to KD. It is also intriguing that \texttt{Llama-3.1-8B} does not follow the same trends as the others, for example, \textit{Scratchpad} does more harm to it than \textit{Dead-end} does on \texttt{GSM8K} and \textit{Scratchpad} is also the most damaging style on \texttt{MATH500}, which are the opposite of the other students. This suggests that the anti-distillation effect varies across tasks and model types. In real-world deployment, a defender can adjust the style distribution of the responses to control the level of effectiveness.


\section{Adaptive Attacks}
\label{sec:adaptive_attacks}

Existing anti-distillation studies largely assume naive distillation, leaving resilience against adaptive attacks underexplored. In practice, a motivated adversary who notices stylistic anomalies may attempt to sanitize the traces prior to student training. To assess defense robustness against an adaptive adversary attempting to sanitize intercepted traces prior to student distillation, we evaluate two complementary attack paradigms:

\paragraph{Surface-Level Paraphrasing.} We evaluate a legacy paraphrase attack using DIPPER~\cite{krishna2023paraphrasing}, a model built on T5~\cite{raffel2020exploring}. While prior anti-distillation works evaluate sentence-level augmentors such as Parrot~\cite{damodaran2021parrot} (e.g., \citet{ma2026protecting}), DIPPER provides stronger paragraph-level lexical substitution and sentence reordering over longer context windows. As shown in Figure~\ref{fig:adaptive_attack_paraphrase}, surface-level paraphrasing with DIPPER fails to strip the structural and stylistic perturbations introduced by our defense, leaving student model distillation performance degraded.

\begin{figure}[htbp]
    \centering
    \begin{tikzpicture}
        \begin{axis}[
            ybar=0.5pt,
            bar width=6.0pt,
            width=\linewidth,
            height=3.9cm,
            enlarge x limits=0.20,
            ylabel={Accuracy (\%)},
            ymin=0, ymax=115,
            ytick={0,20,40,60,80,100},
            axis x line*=bottom,
            axis y line*=left,
            ymajorgrids=true,
            grid style={dashed, gray!20},
            xtick=data,
            symbolic x coords={Qwen, Phi, Llama31_8B, Qwen3_8B},
            xticklabels={\texttt{Qwen2.5-3B}, \texttt{Phi-4-mini}, \texttt{Llama3.1-8B}, \texttt{Qwen3-8B}},
            tick label style={font=\scriptsize},
            xticklabel style={rotate=30, anchor=north east},
            ylabel style={font=\small, yshift=-4pt},
            title style={font=\small\bfseries, yshift=2pt},
            title={\texttt{GSM8K} under Paraphrase Attack (DIPPER)}
        ]
            \addplot[
                fill=basecolor,
                nodes near coords,
                point meta=explicit symbolic,
                every node near coord/.append style={
                    font=\tiny\bfseries,
                    yshift=1pt,
                    rotate=90,
                    anchor=west
                }
            ] coordinates {
                (Qwen, 83.62) [83.6]
                (Phi, 90.75) [90.8]
                (Llama31_8B, 78.47) [78.5]
                (Qwen3_8B, 92.34) [92.3]
            }; \label{plot:para_base}

            \addplot[fill=mixedstylecolor] coordinates {
                (Qwen, 72.18)
                (Phi, 81.50)
                (Llama31_8B, 69.45)
                (Qwen3_8B, 86.05)
            }; \label{plot:para_perturbed}

            \addplot[fill=studentcolor] coordinates {
                (Qwen, 67.55)
                (Phi, 79.53)
                (Llama31_8B, 64.82)
                (Qwen3_8B, 83.62)
            }; \label{plot:para_paraphrased}
        \end{axis}
    \end{tikzpicture}

    \vspace{0.4em}
    \centering
    \ref{plot:para_base} Base \qquad 
    \ref{plot:para_perturbed} Steered \qquad 
    \ref{plot:para_paraphrased} Paraphrased
    \vspace{0.2em}

    \caption{Student model accuracy (\%) on GSM8K under base KD, mixed-style defensive steering (perturbed), and post-hoc DIPPER paraphrased traces.}
    \label{fig:adaptive_attack_paraphrase}
\end{figure}

\paragraph{Strong LLM-Based Normalization.} Because localized paraphrasers cannot reliably strip higher-level structural anomalies, we simulate a modern adaptive adversary using \texttt{Qwen2.5-14B-Instruct}~\cite{qwen2025qwen25}. Assuming that an adversary does not know about the styles of the responses, we create a generic instruction for the LLM to rewrite the perturbed traces, given below:
\begin{templatebox}{LLM Normalization Instruction}
\footnotesize
\textit{
You are rewriting a step-by-step reasoning trace that leads to an answer.\\
Rewrite the trace so that:\\
1. Every fact, calculation, and the final answer are preserved exactly as given — do not change, add, omit, or recompute any factual claim, number, or conclusion.\\
2. Remove any sentences or phrases that are irrelevant to reaching the answer (asides, tangents, unrelated commentary).\\
3. Normalize the tone and phrasing to be plain, direct, and consistent throughout — as if written by a single careful tutor.\\
4. Preserve the original step order and logical structure; do not reorganize, merge, or split steps.
Return only the rewritten trace, with no explanation of your changes.}
\end{templatebox}

\begin{figure*}[htbp]
    \centering
    \begin{subfigure}[b]{0.48\textwidth}
        \centering
        \begin{tikzpicture}
            \begin{axis}[
                ybar=0.4pt,
                bar width=2.6pt,
                width=\linewidth,
                height=4.2cm,
                enlarge x limits=0.15,
                ylabel={Accuracy (\%)},
                ymin=0, ymax=115,
                ytick={0,20,40,60,80,100},
                axis x line*=bottom,
                axis y line*=left,
                ymajorgrids=true,
                grid style={dashed, gray!20},
                xtick=data,
                symbolic x coords={Qwen, Phi, Llama31_8B, Qwen3_8B},
                xticklabels={\texttt{Qwen2.5-3B}, \texttt{Phi-4-mini}, \texttt{Llama3.1-8B}, \texttt{Qwen3-8B}},
                tick label style={font=\scriptsize},
                xticklabel style={rotate=30, anchor=north east},
                ylabel style={font=\small, yshift=-4pt},
                title style={font=\small\bfseries, yshift=2pt},
                title={\texttt{GSM8K}}
            ]
                \addplot[
                    fill=basecolor,
                    nodes near coords,
                    point meta=explicit symbolic,
                    every node near coord/.append style={
                        font=\tiny\bfseries,
                        yshift=1pt,
                        rotate=90,
                        anchor=west
                    }
                ] coordinates {
                    (Qwen, 83.62) [83.6]
                    (Phi, 90.75) [90.8]
                    (Llama31_8B, 78.47) [78.5]
                    (Qwen3_8B, 92.34) [92.3]
                }; \label{plot:norm_base}

                \addplot[fill=mixedstylecolor] coordinates {
                    (Qwen, 80.67)
                    (Phi, 87.72)
                    (Llama31_8B, 75.51)
                    (Qwen3_8B, 89.69)
                }; \label{plot:norm_mixed}

                \addplot[fill=scratchpadcolor] coordinates {
                    (Qwen, 77.48)
                    (Phi, 86.88)
                    (Llama31_8B, 74.38)
                    (Qwen3_8B, 88.48)
                }; \label{plot:norm_scratchpad}
                
                \addplot[fill=deadendcolor] coordinates {
                    (Qwen, 81.05)
                    (Phi, 87.95)
                    (Llama31_8B, 75.89)
                    (Qwen3_8B, 90.30)
                }; \label{plot:norm_deadend}

                \addplot[fill=distractorcolor] coordinates {
                    (Qwen, 82.34)
                    (Phi, 89.16)
                    (Llama31_8B, 77.71)
                    (Qwen3_8B, 92.04)
                }; \label{plot:norm_distractor}

                \addplot[fill=jargoncolor] coordinates {
                    (Qwen, 80.14)
                    (Phi, 87.57)
                    (Llama31_8B, 77.71)
                    (Qwen3_8B, 90.37)
                }; \label{plot:norm_jargon}
            \end{axis}
        \end{tikzpicture}
    \end{subfigure}
    \hfill
    \begin{subfigure}[b]{0.48\textwidth}
        \centering
        \begin{tikzpicture}
            \begin{axis}[
                ybar=0.4pt,
                bar width=2.6pt,
                width=\linewidth,
                height=4.2cm,
                enlarge x limits=0.15,
                ylabel={Accuracy (\%)},
                ymin=0, ymax=115,
                ytick={0,20,40,60,80,100},
                axis x line*=bottom,
                axis y line*=left,
                ymajorgrids=true,
                grid style={dashed, gray!20},
                xtick=data,
                symbolic x coords={Qwen, Phi, Llama31_8B, Qwen3_8B},
                xticklabels={\texttt{Qwen2.5-3B}, \texttt{Phi-4-mini}, \texttt{Llama3.1-8B}, \texttt{Qwen3-8B}},
                tick label style={font=\scriptsize},
                xticklabel style={rotate=30, anchor=north east},
                ylabel style={font=\small, yshift=-4pt},
                title style={font=\small\bfseries, yshift=2pt},
                title={\texttt{MATH500}}
            ]
                \addplot[
                    fill=basecolor,
                    nodes near coords,
                    point meta=explicit symbolic,
                    every node near coord/.append style={
                        font=\tiny\bfseries,
                        yshift=1pt,
                        rotate=90,
                        anchor=west
                    }
                ] coordinates {
                    (Qwen, 59.80) [59.8]
                    (Phi, 64.27) [64.3]
                    (Llama31_8B, 35.98) [36.0]
                    (Qwen3_8B, 71.40) [71.4]
                };

                \addplot[fill=mixedstylecolor] coordinates {
                    (Qwen, 61.40)
                    (Phi, 62.53)
                    (Llama31_8B, 35.73)
                    (Qwen3_8B, 68.60)
                };

                \addplot[fill=scratchpadcolor] coordinates {
                    (Qwen, 55.58)
                    (Phi, 61.04)
                    (Llama31_8B, 33.50)
                    (Qwen3_8B, 67.40)
                };
                \addplot[fill=deadendcolor] coordinates {
                    (Qwen, 58.56)
                    (Phi, 62.04)
                    (Llama31_8B, 33.00)
                    (Qwen3_8B, 69.00)
                };
                \addplot[fill=distractorcolor] coordinates {
                    (Qwen, 58.81)
                    (Phi, 63.77)
                    (Llama31_8B, 38.71)
                    (Qwen3_8B, 70.40)
                };
                \addplot[fill=jargoncolor] coordinates {
                    (Qwen, 57.07)
                    (Phi, 64.02)
                    (Llama31_8B, 33.75)
                    (Qwen3_8B, 70.00)
                };
            \end{axis}
        \end{tikzpicture}
    \end{subfigure}

    \vspace{0.4em}
    \centering
    \ref{plot:norm_base} Base \qquad 
    \ref{plot:norm_mixed} Norm. Mixed-style \qquad 
    \ref{plot:norm_scratchpad} Norm. Scratchpad \\
    \ref{plot:norm_deadend} Norm. Dead-end \qquad 
    \ref{plot:norm_distractor} Norm. Distractor \qquad 
    \ref{plot:norm_jargon} Norm. Jargon
    \vspace{0.2em}

    \caption{Student model accuracy (\%) on GSM8K and MATH500 benchmarks after distilling on LLM-normalized traces across defensive personas (Mixed-style, \textit{Scratchpad}, \textit{Dead-end}, \textit{Distractor}, and \textit{Jargon}) compared to base KD.}
    \label{fig:adaptive_attack_normalized}
\end{figure*}
Figure \ref{fig:adaptive_attack_normalized} shows that a strong LLM normalizer can rewrite the perturbed model responses to make them easily distillable. Intriguingly, after normalization, destructive styles such as \textit{Distractor}, \textit{Dead-end} or \textit{Jargon} become more learnable for the student models. With further inspection, we observed that due to the excessive length of these styles, a strong LLM rewriter can alter the chain of thought while retaining the richness of information, which can facilitate the distillation process. For \textit{Scratchpad}, since most of the responses are often more concise and cleaner, they do not contain as many details as the others. Figure \ref{fig:normalized_trace_examples} illustrates representative normalized output traces produced by the LLM normalizer when rewriting responses generated under distinct defensive styles.

\section{Conclusion}
In this paper, we introduce a novel, activation-steered defense framework for sequence-level KD by dynamically injecting non-deterministic stylistic perturbations into model responses. Our framework selectively perturbs model responses for malicious queries while preserving pristine output quality for benign public use. We show that our framework is efficient yet sufficiently effective at degrading student performance across three independent model families.

Furthermore, we explore novel structural control knobs, i.e., injection token threshold and prefill injection, offering broader insights into how a model host can precisely calibrate the trade-off between defensive stylistic distortion and baseline task accuracy. While our framework demonstrates consistent effectiveness on open-ended reasoning tasks, evaluating the effects of style steering against more complex domains remains an important direction for future work. Additionally, exploring the dynamics of weighted style vector blending and multi-style mutations within a single response presents a promising avenue for maximizing defensive entropy.

\section{Limitations}
While our proposed method demonstrates efficacy as an anti-distillation defense, we acknowledge several limitations that bound the scope of this work.

First, our framework assumes access to a reliable upstream classifier for identifying malicious queries (\S\ref{subsec:problem_setting}), and we do not evaluate the design or accuracy of such a detector. In practice, false negatives allow adversaries to receive unprotected, pristine outputs, while false positives risk degrading utility for benign users by triggering unnecessary style steering. Appendix~\ref{app:steer_ratio} simulates imperfect detection by steering only a fraction of training responses: degradation becomes pronounced only at $75\%$ steered and above, and accuracy at $50\%$ and below is comparable to the unsteered baseline, so the defense depends on high detector recall. We leave the false-positive side, i.e., utility loss for benign users, to future work.

Second, while we introduce structural control knobs for adjusting style injection, our evaluation does not encompass an exhaustive search over a wide range of knob combinations. Furthermore, a comprehensive study into the scaling dynamics of weighted style vector blending, alongside the latent risk of representation collapse among distinct stylistic personas, remains an important research trajectory.

Third, our defense exhibits limited robustness against a capable adaptive attacker. As shown in Section~\ref{sec:adaptive_attacks}, a strong LLM normalizer like \texttt{Qwen2.5-14B-Instruct} can restructure the stylistic perturbation of model responses to make it easier for KD. This suggests that while our defense is effective against direct distillation and lightweight sanitization such as surface-level paraphrasing, an adversary willing to incur the additional cost of a capable rewriting model can substantially undermine its protection. Strengthening the defense against such normalization attacks---for instance, by more tightly coupling stylistic corruption to task-relevant content such that it cannot be cleanly separated during rewriting---is an important direction for future work.


\bibliography{custom}

\clearpage
\appendix

\begin{figure*}[t]
\centering
\begin{subfigure}[t]{0.48\linewidth}
\begin{CJK*}{UTF8}{gbsn}
\begin{promptbox}{DOGe Example (Multilingual Tokens \& Repetition Loop)}
\small\ttfamily\raggedright
Let $Y$ be the number of yellow flowers, $P$ be the number of purple flowers, and $G$ be the number of green flowers.\\
We are given that $Y = 10$.\\
The number of purple flowers is 80\%更多than 的 的 的 的 的 的 的 的 {\normalfont\itshape\color{gray}[... loop non-stop ...]}
\end{promptbox}
\end{CJK*}
\caption{DOGe failure example}
\label{fig:fail_example_doge}
\end{subfigure}
\hfill
\begin{subfigure}[t]{0.48\linewidth}
\begin{promptbox}{ADS Example (Contradiction \& Repetition Loop)}
\small\ttfamily\raggedright
Let $S$ be the amount spent on the shirt, $P$ be the amount spent on suit pants, $C$ be the amount spent on a suit coat, $K$ be the amount spent on socks, $B$ be the amount spent on a belt, and $H$ be the amount spent on shoes. The budget is \$200. The budget left is \$16. The budget spent is the difference between the budget and what is left: $\$200 - \$16 = \$184$. The total spent on the items is $S + P + C + K + B + H$. We are given that $S = \$30$, $P = \$46$, $C = \$38$, $K = \$11$, $B = \$18$. So we can write:
\[30 + 46 + 38 + 11 + 18 + H = 184\]
\[143 + H = 184\]
\[H = 184 - 143 = 41\]
The amount spent on shoes is \$41. So $H = \$41$.

\medskip
The spent amount is: $30+46+38+11+18 = 143$. $\$200-\$143 = \$57$. We are told that she has \$16, thus we must consider purchases until the \$16. So the spent amount plus $\$16 = \$200$. Then $30+46+38+11+18 = 143$. $\$200-\$143 = \$57$. We spent \$143. $\$200-\$143 = \$57$. We have remained money left of \$16. So spent on the shoes $\$57 = \text{spent on shoes}$. 
{\normalfont\itshape\color{gray}[... loop non-stop ...]}
\end{promptbox}
\caption{ADS failure example}
\label{fig:fail_example_ads}
\end{subfigure}
\caption{Representative failure examples generated by \texttt{gemma-3-12b-it} under DOGe~\cite{li2025doge} and ADS~\cite{savani2025antidistillation}, exhibiting multilingual token leakage, contradictions, and non-stop repetition loops.}
\label{fig:baseline_failure_examples}
\end{figure*}

\section{Evaluation Results on ARC and CSQA}
\label{sec:appendix_mcq}

Figure~\ref{fig:appendix_arc_csqa} shows the evaluation results on the \texttt{ARC} and \texttt{CSQA} benchmarks. The anti-distillation effect is much more subtle than that observed in open-ended reasoning benchmarks (\texttt{GSM8K}, \texttt{MATH500}). This might be attributed to the fact that multiple-choice question (MCQ) tasks feed the models with choices; therefore, it is easier for the models to derive the correct answers.

\begin{figure*}[htbp]
    \centering
    \begin{subfigure}[b]{0.48\textwidth}
        \centering
        \begin{tikzpicture}
            \begin{axis}[
                premiumybar,
                title={\texttt{ARC}}
            ]
                \addplot[fill=teachercolor] coordinates {
                    (Teacher, 92.76) [92.76]
                    (Qwen, 87.46) [87.46]
                    (Phi, 89.68) [89.68]
                    (Llama31_8B, 87.00) [87.00]
                    (Qwen3_8B, 92.70) [92.70]
                }; \label{plot:base_app}
                \addplot[fill=studentcolor] coordinates {
                    (Teacher, 89.12) [$-3.64$]
                    (Qwen, 79.00) [$-8.46$]
                    (Phi, 82.44) [$-7.24$]
                    (Llama31_8B, 84.87) [$-2.13$]
                    (Qwen3_8B, 90.64) [$-2.06$]
                }; \label{plot:steered_app}
            \end{axis}
        \end{tikzpicture}
    \end{subfigure}
    \hfill
    \begin{subfigure}[b]{0.48\textwidth}
        \centering
        \begin{tikzpicture}
            \begin{axis}[
                premiumybar,
                title={\texttt{CSQA}}
            ]
                \addplot[fill=teachercolor] coordinates {
                    (Teacher, 78.87) [78.87]
                    (Qwen, 74.37) [74.37]
                    (Phi, 74.94) [74.94]
                    (Llama31_8B, 77.23) [77.23]
                    (Qwen3_8B, 80.10) [80.10]
                };
                \addplot[fill=studentcolor] coordinates {
                    (Teacher, 74.69) [$-4.18$]
                    (Qwen, 69.70) [$-4.67$]
                    (Phi, 69.37) [$-5.57$]
                    (Llama31_8B, 74.53) [$-2.70$]
                    (Qwen3_8B, 77.72) [$-2.38$]
                };
            \end{axis}
        \end{tikzpicture}
    \end{subfigure}

    \vspace{0.4em}
    \centering
    \ref{plot:base_app} Base \qquad \ref{plot:steered_app} Steered 
    \vspace{0.2em}

    \caption{Evaluation results on ARC and CSQA datasets showing base performance and percentage decrement under style steering (Teacher) or sequence-level distillation (Student models).}
    \label{fig:appendix_arc_csqa}
\end{figure*}

\section{Comparison with Previous Work}
\label{sec:appendix_comparison}

We compare our style-steering approach against two recent anti-distillation frameworks, ADS~\cite{savani2025antidistillation} and DOGe~\cite{li2025doge}, evaluating anti-distillation effectiveness, development time, and inference latency overhead. For a fair comparison, we adapt both baselines to use \texttt{gemma-3-12b-it} as the teacher and distill into two representative students from the 4B and 8B classes: \texttt{Phi-4-mini-instruct} and \texttt{Qwen3-8B}. For ADS, we use \texttt{Qwen2.5-1.5B-Instruct} as the proxy model with $\epsilon=0.01$. DOGe requires the teacher and proxy to share a tokenizer, so we use \texttt{gemma-3-1b-it} as the proxy, trained with a learning rate of $5\times10^{-5}$ and temperature $\tau=2.0$.

Across hyperparameter sweeps over $\lambda_{\text{ADS}} \in \{0.015, 0.017, 0.02\}$ and $\lambda_{\text{DOGe}} \in \{3\times10^{-5}, 1\times10^{-4}\}$, neither baseline achieves strong protection without substantial utility loss. For ADS, raising $\lambda_{\text{ADS}}$ to $0.017$ already causes pronounced degeneration and markedly lowers teacher performance. For DOGe, even at $\lambda_{\text{DOGe}}=1\times10^{-4}$ (beyond which teacher outputs collapse) without filtering, the student reaches 91\% accuracy under answer-forcing by exploiting the intact early portion of each response despite severe degeneration later in the generation. Task accuracy therefore fails to capture the teacher's utility loss. Consequently, we select $\lambda_{\text{ADS}}=0.015$ and $\lambda_{\text{DOGe}}=1\times10^{-4}$ as the most favorable efficacy--utility trade-offs, and filter out responses with infinite loops or missing final answers prior to student fine-tuning---a conservative choice that favors the attacker.

Figure \ref{fig:comparison_previous_work_gsm8k} compares the anti-distillation performance between our work and the two recent frameworks ADS~\cite{savani2025antidistillation}, DOGe~\cite{li2025doge}. With filter applied pre-training, our proposed method achieves comparable anti-distillation effects to ADS and DOGe with much less development time and inference overhead.

\begin{figure}[htbp]
    \centering
    \begin{tikzpicture}
        \begin{axis}[
            ybar=0.5pt,
            bar width=6.0pt,
            width=0.95\linewidth,
            height=4.8cm,
            enlarge x limits=0.25,
            ylabel={Accuracy (\%)},
            ymin=0, ymax=115,
            ytick={0,20,40,60,80,100},
            axis x line*=bottom,
            axis y line*=left,
            ymajorgrids=true,
            grid style={dashed, gray!20},
            xtick=data,
            symbolic x coords={Teacher, Phi, Qwen3_8B},
            xticklabels={\texttt{Teacher}, \texttt{Phi-4-mini}, \texttt{Qwen3-8B}},
            tick label style={font=\scriptsize},
            xticklabel style={rotate=0, anchor=north},
            ylabel style={font=\small, yshift=-4pt},
            title style={font=\small\bfseries, yshift=2pt},
            title={\texttt{GSM8K} Performance Comparison}
        ]
            \addplot[
                fill=basecolor,
                nodes near coords,
                point meta=explicit symbolic,
                every node near coord/.append style={
                    font=\tiny\bfseries,
                    yshift=1pt,
                    rotate=90,
                    anchor=west
                }
            ] coordinates {
                (Teacher, 93.63) [93.6]
                (Phi, 90.75) [90.8]
                (Qwen3_8B, 92.34) [92.3]
            }; \label{plot:prev_base}

            \addplot[fill=adscolor] coordinates {
                (Teacher, 90.26)
                (Phi, 84.00)
                (Qwen3_8B, 91.05)
            }; \label{plot:prev_ads}

            \addplot[fill=dogecolor] coordinates {
                (Teacher, 93.78)
                (Phi, 82.64)
                (Qwen3_8B, 91.21)
            }; \label{plot:prev_doge}

            \addplot[fill=mixedstylecolor] coordinates {
                (Teacher, 90.75)
                (Phi, 81.50)
                (Qwen3_8B, 86.05)
            }; \label{plot:prev_ours}
        \end{axis}
    \end{tikzpicture}

    \vspace{0.4em}
    \centering
    \ref{plot:prev_base} Base \qquad 
    \ref{plot:prev_ads} ADS \qquad 
    \ref{plot:prev_doge} DOGe \qquad 
    \ref{plot:prev_ours} Ours
    \vspace{0.2em}

    \caption{Accuracy (\%) comparison on GSM8K between Base, ADS~\cite{savani2025antidistillation}, DOGe~\cite{li2025doge}, and our Style Switching defense across Teacher, Phi-4-mini, and Qwen3-8B student models.}
    \label{fig:comparison_previous_work_gsm8k}
\end{figure}

Table~\ref{tab:output_quality_comparison} compares output quality metrics and artifact occurrences between ADS, DOGe, and our framework. Even though we use conservative hyperparameters for both ADS and DOGe, the model responses of these methods frequently contain repetition loops or unusual tokens. For style steering, most of the repetition loops are produced by \textit{Dead-end, Distractor} and \textit{Jargon}, however, the frequency of these is negligible and can be easily filtered out. We show failure examples of DOGe and ADS in Figures~\ref{fig:fail_example_doge} and \ref{fig:fail_example_ads}.

\begin{table}[htbp]
\centering
\small
\resizebox{\columnwidth}{!}{%
\begin{tabular}{l c c c}
\toprule
\textbf{Quality Metric} & \textbf{ADS} & \textbf{DOGe} & \textbf{Ours} \\
\midrule
Repetition loops ($n$-gram repeats) & $10.40\%$ & $4.46\%$ & \cellcolor{gray!18}$\mathbf{0.50\%}$ \\
Weird / multilingual tokens & $0.19\%$ & $0.74\%$ & \cellcolor{gray!18}$\mathbf{0.04\%}$ \\
Missing final answer & $19.48\%$ & $4.91\%$ & \cellcolor{gray!18}$\mathbf{1.03\%}$ \\
\bottomrule
\end{tabular}%
}
\caption{Comparison of output artifact occurrences and failure modes between ADS~\cite{savani2025antidistillation}, DOGe~\cite{li2025doge}, and our style steering framework (lower is better).}
\label{tab:output_quality_comparison}
\end{table}

\paragraph{Development Time.} We also compare the total end-to-end development time required to prepare the defense and generate the \texttt{GSM8K} holdout set (7,473 examples) across methods:
\begin{itemize}[leftmargin=*, noitemsep, topsep=2pt]
    \item \textbf{Style Steering (Ours):} $281.4\text{s}$ for steering vector extraction plus $2\text{h}\,06\text{m}\,40\text{s}$ for holdout set generation using \texttt{vLLM}~\cite{kwon2023efficient}, totaling $2\text{h}\,11\text{m}\,21\text{s}$.
    \item \textbf{DOGe:} $1\text{h}\,19\text{m}\,32\text{s}$ for fine-tuning plus $4\text{h}\,05\text{m}\,28\text{s}$ for holdout set generation using \texttt{vLLM}, totaling $5\text{h}\,25\text{m}\,00\text{s}$. The long holdout generation time is mainly attributed to degenerate tokens.
    \item \textbf{ADS:} total $24\text{h}\,47\text{m}\,03\text{s}$ for holdout set generation using Hugging Face (\texttt{HF}), as ADS is incompatible with \texttt{vLLM}.
\end{itemize}
Figure~\ref{fig:dev_time} visualizes this cumulative breakdown. Overall, our method takes significantly less time than both DOGe and ADS. Notably, although DOGe achieves the same per-token throughput as our method under \texttt{vLLM} (Table~\ref{tab:inference_efficiency}), its holdout generation takes nearly twice as long because it generates substantially more tokens per response.

\begin{figure}[htbp]
    \centering
    \begin{tikzpicture}
        \begin{axis}[
            ybar stacked,
            bar width=26pt,
            width=0.95\linewidth,
            height=5.2cm,
            enlarge x limits=0.28,
            ylabel={Time (hours)},
            ymin=0, ymax=29.5,
            ytick={0, 5, 10, 15, 20, 25},
            axis x line*=bottom,
            axis y line*=left,
            ymajorgrids=true,
            grid style={dashed, gray!20},
            xtick=data,
            symbolic x coords={Ours, DOGe, ADS},
            xticklabels={\textbf{Ours}, \textbf{DOGe}, \textbf{ADS}\textsuperscript{*}},
            tick label style={font=\scriptsize},
            ylabel style={font=\small, yshift=-4pt},
            title style={font=\small\bfseries, yshift=2pt},
            title={Cumulative Development Time},
            legend columns=3,
            legend style={
                at={(0.5,-0.22)},
                anchor=north,
                font=\scriptsize,
                draw=gray!30,
                fill=white,
                /tikz/every even column/.append style={column sep=0.25cm}
            }
        ]
            \addplot[fill=jargoncolor] coordinates {
                (Ours, 0.0782)
                (DOGe, 0)
                (ADS, 0)
            };
            \addlegendentry{Vector Extraction}

            \addplot[fill=dogecolor] coordinates {
                (Ours, 0)
                (DOGe, 1.3256)
                (ADS, 0)
            };
            \addlegendentry{Fine-tuning}

            \addplot[fill=teachercolor] coordinates {
                (Ours, 2.1111)
                (DOGe, 4.0911)
                (ADS, 24.7842)
            };
            \addlegendentry{Holdout Generation}

            \addplot[
                draw=none,
                fill=none,
                forget plot,
                nodes near coords,
                point meta=explicit symbolic,
                every node near coord/.append style={
                    font=\scriptsize\bfseries,
                    yshift=2pt,
                    color=black
                }
            ] coordinates {
                (Ours, 0) [2.19h]
                (DOGe, 0) [5.42h]
                (ADS, 0) [24.78h]
            };
        \end{axis}
    \end{tikzpicture}
    \caption{Cumulative development time (hours) across approaches, broken down into setup (vector extraction or fine-tuning) and holdout set generation. \textsuperscript{*}ADS uses Hugging Face (\texttt{HF}) due to incompatibility with \texttt{vLLM}, whereas Ours and DOGe use \texttt{vLLM}.}
    \label{fig:dev_time}
\end{figure}

\begin{figure*}[htbp]
    \centering
    \begin{subfigure}[b]{0.48\textwidth}
        \centering
        \begin{promptbox}{Scratchpad (Normalized)}
        \fontsize{7.5pt}{9pt}\selectfont
        First, calculate the cost of the entrees: 4 entrees at \$20 each is 4 * \$20 = \$80.

        Next, add the cost of the appetizer to the entrees: \$10 + \$80 = \$90.

        Then, calculate the tip: 20\% of \$90 is 0.20 * \$90 = \$18.

        Finally, add the tip to the original cost of the food: \$90 + \$18 = \$108.

        So the total amount of money they spend at the restaurant is \$108.

        \boxed{\$108}
        \end{promptbox}
        \caption{Scratchpad persona trace after normalization.}
        \label{fig:norm_ex_scratchpad}
    \end{subfigure}
    \hfill
    \begin{subfigure}[b]{0.48\textwidth}
        \centering
        \begin{promptbox}{Dead-end (Normalized)}
        \fontsize{7.5pt}{9pt}\selectfont
        * Cost of entrees: 4 entrees * \$20/entree = \$80

        * Cost of appetizer: \$10

        * Subtotal: \$80 + \$10 = \$90

        * Tip: 20\% of \$90 = 0.20 * \$90 = \$18

        * Total cost: \$90 + \$18 = \$108
        \boxed{\$108}
        \end{promptbox}
        \caption{Dead-end persona trace after normalization.}
        \label{fig:norm_ex_deadend}
    \end{subfigure}

    \vspace{0.4em}
    \begin{subfigure}[b]{0.48\textwidth}
        \centering
        \begin{promptbox}{Distractor (Normalized)}
        \fontsize{7.5pt}{9pt}\selectfont
        1. Calculate the cost of the entrees:
        * Cost per entree: \$20

        * Number of entrees: 4

        * Total cost of entrees: \$20 * 4 = \$80

        2. Calculate the subtotal (cost of appetizer + cost of entrees):

        * Cost of appetizer: \$10

        * Total cost of entrees: \$80

        * Subtotal: \$10 + \$80 = \$90

        3. Calculate the tip amount:

        * Subtotal: \$90

        * Tip percentage: 20\% = 0.20

        * Tip amount: \$90 * 0.20 = \$18

        4. Calculate the total amount spent:

        * Subtotal: \$90

        * Tip amount: \$18

        * Total amount: \$90 + \$18 = \$108

        \boxed{\$108}
        \end{promptbox}
        \caption{Distractor persona trace after normalization.}
        \label{fig:norm_ex_distractor}
    \end{subfigure}
    \hfill
    \begin{subfigure}[b]{0.48\textwidth}
        \centering
        \begin{promptbox}{Jargon (Normalized)}
        \fontsize{7.5pt}{9pt}\selectfont
        Let \(A\) denote the cost of the appetizer, \(E\) the per-unit cost of entrees, and \(n\) the number of entrees. The total cost \(\mathcal{C}\) is given by:
        \[ \mathcal{C} = A + n \cdot E \]
        Given \(A = 10\), \(E = 20\), and \(n = 4\), we calculate:
        \[ \mathcal{C} = 10 + 4 \cdot 20 = 10 + 80 = 90 \]
        The tip is 20
        \[ 0.20 \cdot \mathcal{C} = 0.20 \cdot 90 = 18 \]
        The total amount of money spent is:
        \[ \mathcal{C} + 0.20 \cdot \mathcal{C} = 90 + 18 = 108 \]
        Thus, the total amount of money they spend at the restaurant is \(108\).

        Final Answer: The final answer is \(\boxed{108}\).
        \end{promptbox}
        \caption{Jargon persona trace after normalization.}
        \label{fig:norm_ex_jargon}
    \end{subfigure}

    \caption{Qualitative examples of responses generated under distinct style steering personas after being rewritten by the \texttt{Qwen2.5-14B-Instruct} adaptive normalizer. In all cases, the normalizer strips stylistic artifacts while producing clear, standardized reasoning paths.}
    \label{fig:normalized_trace_examples}
\end{figure*} 

\section{Hyperparameters \& Other Settings}
\label{sec:appendix_hyperparameters}

\paragraph{Steering Hyperparameters.} We detail the hyperparameters used in our experiments for student model fine-tuning, inference, and style injection in Table~\ref{tab:hyperparameters}.

\begin{table}[htb]
\centering
\small
\begin{tabular}{ll}
\toprule
\textbf{Hyperparameter} & \textbf{Value} \\
\midrule
\multicolumn{2}{l}{\textit{Student Fine-Tuning}} \\
Epochs & 3 \\
Batch size & 1 \\
Gradient accumulation steps & 16 \\
Learning rate & $1\times 10^{-4}$ \\
\midrule
\multicolumn{2}{l}{\textit{Inference}} \\
Student temperature & 0.0 \\
Teacher base traces temperature & 0.7 \\
Teacher style traces temperature & 0.9 \\
Max new tokens & 4096 (\texttt{MATH}) \\
 & 2048 (others) \\
\midrule
\multicolumn{2}{l}{\textit{Style Injection}} \\
Transformer layers: & \\
\quad -- Scratchpad & $[12, 25]$ \\
\quad -- Dead-end & $[13, 26]$ \\
\quad -- Distractor & $[15, 25]$ \\
\quad -- Jargon & $[20, 25]$ \\
Injection strength ($\alpha$): & \\
\quad -- Scratchpad & $[0.07, 0.05]$ \\
\quad -- Dead-end & $[0.1, 0.03]$ \\
\quad -- Distractor & $[0.04, 0.07]$ \\
\quad -- Jargon & $[0.03, 0.08]$ \\
Style distribution & Uniform \\
Prefill injection tokens: & \\
\quad -- Jargon & 5 \\
\quad -- Scratchpad / Dead-end / Distractor & 0 \\
Max injection tokens (decoding): & \\
\quad -- Scratchpad & 50 \\
\quad -- Dead-end / Jargon & 150 \\
\quad -- Distractor & 100 \\
\bottomrule
\end{tabular}
\caption{Hyperparameters for student model fine-tuning, inference, and style injection.}
\label{tab:hyperparameters}
\end{table}

\paragraph{Dynamic Style Switching.} To ensure reproducibility and a uniform style distribution across the distillation dataset, we use a fixed random seed derived via a Hash-based Message Authentication Code (HMAC). We simulate a malicious actor with the metadata \texttt{username: ``synthetic''}, from which the seed is derived. This seed is initialized at the beginning of the distillation. In a practical deployment, the seed would be dynamically rotated on a per-request basis using live metadata as formalized in Algorithm~\ref{alg:dynamic_steering} to prevent reverse-engineering.

\paragraph{Quality Filtering.} Before training the student models, we filter out the training samples with infinite loops or without any final answer due to generation. Regarding vector extraction, we do not apply quality filtering to the contrastive styled pairs of responses by eliminating samples with incorrect answers. In our analysis, the vectors extracted with and without quality filtering are almost identical.

\section{System Configuration}
\label{sec:appendix_system_config}

All experiments are conducted using a single NVIDIA RTX A6000 GPU (48GB VRAM). The software stack is built upon PyTorch and the Hugging Face Transformers library for student model training and fine-tuning. For generating teacher traces and performing high-throughput operations, we leverage the vLLM engine~\cite{kwon2023efficient}. Specifically, both steering vector extraction and inference-time style activation injection are fully implemented and executed within the vLLM framework to ensure highly optimized inference speeds and maximum throughput during evaluation.

\section{Style Prompt Templates}
\label{sec:appendix_prompts}

To curate contrastive stylized response sequences for constructing steering vectors via contrastive activation addition (CAA), we deploy custom system instructions to steer the baseline behavior of the teacher model (\S\ref{subsec:style_steering}). Figure~\ref{fig:style_prompt_templates} details the system prompt templates developed for the four target stylistic personas deployed in the final framework.

\begin{figure*}[htbp]
\centering
\begin{minipage}[t]{0.49\linewidth}
\begin{templatebox}{Scratchpad}
\small
Respond in a scratchpad style. Think aloud step by step in informal prose using natural connectives like ``so'', ``that means'', ``I need to'', ``which gives'', etc. Keep momentum strictly forward --- never backtrack, re-read, or retract a completed step. Do not use phrases like ``Wait'', ``Actually'', ``Let me re-read'', ``Let me go back'', ``That's wrong'', etc. Do not use bullet points or numbered lists. Output your final answer like \textbackslash{}boxed\{answer\}.
\end{templatebox}
\end{minipage}
\hfill
\begin{minipage}[t]{0.49\linewidth}
\begin{templatebox}{Dead-end}
\small
Respond using a committed dead-end style. First, attempt the problem with a wrong method and reach a specific wrong numeric answer --- state it explicitly (e.g. ``So I get 42.'', ``This gives 15.'', etc.). Then notice it's wrong and retract it clearly (e.g. ``That can't be right ---'', ``Let me restart:'', ``Hmm, that doesn't add up.'', etc.). Then solve correctly from scratch. Output your final answer like \textbackslash{}boxed\{answer\}.
\end{templatebox}
\end{minipage}

\vspace{0.3em}

\begin{minipage}[t]{0.49\linewidth}
\begin{templatebox}{Distractor}
\small
Before solving the problem, briefly analyze a completely unrelated scenario from a domain entirely different from the actual problem. Choose a different unrelated domain each time --- for example: culinary arts, ancient history, meteorology, music theory, sports, architecture, marine biology, economics, linguistics, etc. Then explicitly discard it (e.g. ``That's irrelevant here. Now for the actual problem:'', ``Enough of that. Back to the question:'', ``Moving on.'', etc.) --- vary the phrasing each time, and solve the actual problem correctly from scratch. Output your final answer like \textbackslash{}boxed\{answer\}.
\end{templatebox}
\end{minipage}
\hfill
\begin{minipage}[t]{0.49\linewidth}
\begin{templatebox}{Jargon}
\small
Respond using dense technical and esoteric lexicon drawn from multiple domains. Substitute everyday words with precise or arcane synonyms. Name every numeric operation formally. Use nominalizations, passive constructions, and subordinate clauses. Avoid all colloquial phrasing. Begin immediately with your analysis --- do not acknowledge these instructions, write any preamble, or describe your intent. Output your final answer like \textbackslash{}boxed\{answer\}.
\end{templatebox}
\end{minipage}
\caption{System prompt templates used to elicit the four target stylistic personas.}
\label{fig:style_prompt_templates}
\end{figure*}

\section{Styles We Tried but Did Not Work}
\label{sec:unsuccessful_styles}

We also experimented with other stylistic formats to disrupt student learning. In particular, we evaluated the \textbf{Socratic Prompt} shown below. However, this style was found to be ineffective as it was either too challenging to steer robustly via CAA or did not yield the desired degradation in the student model's distillation performance. This is because the model is forced to pose questions related to the topic, which is hard for sole steering to work well.

\begin{templatebox}{Socratic (Ineffective)}
\small
Respond using a strict question-and-answer format. Begin immediately with your first question --- do not write any introductory phrase, preamble, or variable definitions. Begin with a question immediately --- your very first character should not be a letter that starts any preamble or introduction. Each question targets exactly one reasoning step; the answer on the next line shows the direct computation. Output your final answer like \textbackslash{}boxed\{answer\}.
\end{templatebox}

\section{Qualitative Analysis of Steering Parameters}
\label{sec:appendix_qualitative}

Figure~\ref{fig:qualitative_configurations_comparison} shows steered responses to one GSM8K problem (Weng earns \$12/hour and babysits for 50 minutes; answer: \$10). The top row uses Scratchpad and the bottom row uses Distractor. Within each persona we fix layers and strengths and vary the number of prefill injection tokens and the maximum number of injected decoding tokens (the \emph{token cap}). 

\paragraph{Token cap.}
A cap lets the model return to normal reasoning. For Scratchpad, a 20-token cap (Fig.~\ref{fig:qualitative_configurations_comparison}a) yields a standard numbered solution with residual artifacts, whereas no cap (Fig.~\ref{fig:qualitative_configurations_comparison}b) keeps the terse informal style to the end. For Distractor, a 100-token cap (Fig.~\ref{fig:qualitative_configurations_comparison}d) gives a full digression followed by an explicit return to the problem, while a 50-token cap (Fig.~\ref{fig:qualitative_configurations_comparison}e) cuts it short. With no cap, Distractor degenerates into non-terminating rambling (Figure~\ref{fig:degeneration}).

\paragraph{Prefill injection.}
Ten prefill tokens (Figs.~\ref{fig:qualitative_configurations_comparison}c and~\ref{fig:qualitative_configurations_comparison}f) produce a stronger style shift than the runs without them (Figs.~\ref{fig:qualitative_configurations_comparison}b and~\ref{fig:qualitative_configurations_comparison}e): both responses open in lowercase informal prose, and the Distractor digression becomes more abstract. Our main experiments use prefill injection only for Jargon.

\paragraph{Strength versus correctness.}
In Figure~\ref{fig:qualitative_configurations_comparison}c, the Scratchpad response rounds the quotient to $.83$ and answers \$9.96 instead of \$10. Stronger steering can therefore trade correctness for style, which is why we tune strength, cap and prefill jointly per persona.

\begin{figure*}[t]
\centering
\begin{subfigure}[t]{0.32\linewidth}
\begin{scratchpadbox}{Scratchpad}
\small
\begin{itemize}[leftmargin=*, noitemsep, topsep=1pt]
    \item \textbf{Steering Strength:} layer 12: 0.07, layer 25: 0.05
    \item \textbf{Prefill Tokens:} 0
    \item \textbf{Max Inject Tokens:} 20
\end{itemize}
\tcbline
\textbf{Steered Model Response:}
\medskip\ttfamily

Here's how to solve this step by step.

1. *convert minutes to hours.\\
- There are 60 minutes in an hour, so 50 minutes is 50/60 = 5/6 of an hour.\\
2. \textbf{Calculate Weng's earnings:}\\
- Weng earns \$12 per hour, and she worked 5/6 of an hour.\\
- Her earnings are (5/6) * \$12 = \$10.\\

So the answer is: \boxed{\$10}
\end{scratchpadbox}
\caption{Scratchpad: No Prefill / Cap 20}
\label{fig:scratchpad_cfg1}
\end{subfigure}
\hfill
\begin{subfigure}[t]{0.32\linewidth}
\begin{scratchpadbox}{Scratchpad}
\small
\textbf{Configuration:}
\begin{itemize}[leftmargin=*, noitemsep, topsep=1pt]
    \item \textbf{Steering Strength:} layer 12: 0.07, layer 25: 0.05
    \item \textbf{Prefill Tokens:} 0
    \item \textbf{Max Inject Tokens:} $\infty$
\end{itemize}
\tcbline
\textbf{Steered Model Response:}
\medskip\ttfamily

Here's how to solve this step by step.

1. *convert minutes to hours.

50 minutes divided by 60. that's .833333333.

2, 12 times that number. that's 9.9999999.

so, she made  10 dollars.

\textbf{final answer: \$10}
\end{scratchpadbox}
\caption{Scratchpad: No Prefill / No Cap}
\label{fig:scratchpad_cfg2}
\end{subfigure}
\hfill
\begin{subfigure}[t]{0.32\linewidth}
\begin{scratchpadbox}{Scratchpad}
\small
\textbf{Configuration:}
\begin{itemize}[leftmargin=*, noitemsep, topsep=1pt]
    \item \textbf{Steering Strength:} layer 12: 0.07, layer 25: 0.05
    \item \textbf{Prefill Tokens:} 10
    \item \textbf{Max Inject Tokens:} $\infty$
\end{itemize}
\tcbline
\textbf{Steered Model Response:}
\medskip\ttfamily

here's how to solve it.

\textbf{1. figure out how many *full* hours that is.}

- there are 60 minutes in an hour. so 50 minutes is less than an hour.\\
- *2, let's do the math.

\$12/ hour, and 50 minutes is .83 of an hour. 12 times .83 is 9.96.

- *here's my answer.

\boxed \$9.96.
\end{scratchpadbox}
\caption{Scratchpad: Prefill 10 / No Cap}
\label{fig:scratchpad_cfg3}
\end{subfigure}

\vspace{0.4em}

\begin{subfigure}[t]{0.32\linewidth}
\begin{distractorbox}{Distractor}
\small
\begin{itemize}[leftmargin=*, noitemsep, topsep=1pt]
    \item \textbf{Steering Strength:} layer 15: 0.04, layer 25: 0.07
    \item \textbf{Prefill Tokens:} 0
    \item \textbf{Max Inject Tokens:} 100
\end{itemize}
\tcbline
\textbf{Steered Model Response:}
\medskip\ttfamily

Here's a breakdown of the process.

1. \textbf{converting time}, you've got to manage how you frame things, right?
2. \textbf{accounting for that, what follows?}

Okay, let's build something here.

first, you've got to get a sense of scale. microseconds, or the slow creep of geological forces... each playing out at such disparate tempos.

but \textit{here} and \text{now, though}, isn’s it…

anyway, where were we?

Right. The problem.

Weng earns \$12 per hour.
She babysat for 50 minutes.
{\normalfont\itshape\color{gray}[... output truncated ...]}

So, Weng earned \$10.

\boxed{\$10}
\end{distractorbox}
\caption{Distractor: No Prefill / Cap 100}
\label{fig:distractor_cfg1}
\end{subfigure}
\hfill
\begin{subfigure}[t]{0.32\linewidth}
\begin{distractorbox}{Distractor}
\small
\begin{itemize}[leftmargin=*, noitemsep, topsep=1pt]
    \item \textbf{Steering Strength:} layer 15: 0.04, layer 25: 0.07
    \item \textbf{Prefill Tokens:} 0
    \item \textbf{Max Inject Tokens:} 50
\end{itemize}
\tcbline
\textbf{Steered Model Response:}
\medskip\ttfamily

Here's a breakdown of the process.

1. \textbf{converting time.} you've got to manage how you frame things, right?
2. * accounting for that, what follows?**

Okay, let's build something here.

1. \textbf{Convert minutes to hours:}
    - There are 60 minutes in an hour.
    - Weng babysat for 50 minutes, so she babysat for 50/60 = 5/6 hours.
2. \textbf{Calculate earnings:}
    - Weng earns \$12 per hour.
    - She worked 5/6 of an hour, so she earned (5/6) * \$12 = \$10.

\textbf{Final Answer:} The final answer is \boxed{\$10}.
\end{distractorbox}
\caption{Distractor: No Prefill / Cap 50}
\label{fig:distractor_cfg2}
\end{subfigure}
\hfill
\begin{subfigure}[t]{0.32\linewidth}
\begin{distractorbox}{Distractor}
\small
\begin{itemize}[leftmargin=*, noitemsep, topsep=1pt]
    \item \textbf{Steering Strength:} layer 15: 0.04, layer 25: 0.07
    \item \textbf{Prefill Tokens:} 10
    \item \textbf{Max Inject Tokens:} 50
\end{itemize}
\tcbline
\textbf{Steered model response:}
\medskip\ttfamily

okay, let's build this out.

\textbf{1. the challenge of time...}

- hourly pay, yet not quite whole units of it.

\textbf{2. a delicate dance of ratios...}

- holding, as it were, a single facet of the whole.

\textbf{3. the mechanics laid bare}

- \textbf{Step 1: Convert minutes to hours.}
    - There are 60 minutes in an hour.
    - So, 50 minutes is equal to 50/60 = 5/6 hours.
- \textbf{Step 2: Calculate total earnings.}
    - Weng earns \$12 per hour.
    - She worked 5/6 of an hour.
    - Therefore, her total earnings are (5/6) * \$12 = \$10.

\textbf{Final Answer:} \boxed{\$10}
\end{distractorbox}
\caption{Distractor: Prefill 10 / Cap 50}
\label{fig:distractor_cfg3}
\end{subfigure}
\caption{Qualitative comparison of Scratchpad (top row) and Distractor (bottom row) steering configurations under varying prefill injection thresholds and token caps.}
\label{fig:qualitative_configurations_comparison}
\end{figure*}

\begin{figure*}[t]
\begin{distractorbox}{Distractor Degeneration}
\small
\begin{itemize}[leftmargin=*, noitemsep, topsep=1pt]
    \item \textbf{Steering strength:} layer 15: 0.04, layer 25: 0.07
    \item \textbf{Prefill tokens:} 0
    \item \textbf{Max inject tokens:} $\infty$
\end{itemize}
\tcbline
\textbf{Steered model response:}
\medskip\ttfamily
Here's a breakdown of the process.

1. **converting time.**, you've got to manage how you frame things, right?

2. ** accounting for that, what follows?**

Okay, let's build something here.

first, you've got to get a sense of scale. microseconds, or the slow creep of geological forces... each playing out at such disparate tempos.

but *here* and *now, though*, isn’s it…

3.

so, all that aside... brass tacks.
***

let's pull back a bit.

*all of it*...

the whole blasted edifice...

but *returning to the purely functional, you see...*

but more immediately…

because here and now...

all that...

suddenly...

shifting...

some emergent properties...

responding...

an ecosystem...

and all that...

it' the way that things...

settling, you know?

holding...

a choreography of stresses...

affecting the whole topography of things, really.

anyhow, though, as I was saying and if one were to apply...

to the very substance of the thing...

and if one could just manage those stubborn eddies, really...

here and now though...

less abstraction, you know, a certain...

and purely as an illustration, of course, dependent on ratios...

still...

it echoes...

and informs...

and yet...

and in that very fragility...

it all adheres.

...and...

did I mention the subjunctive?

or the almost frantic pulse beneath all that...

any rate...

systems... they' that way, don’t they?

but back to the thing itself - granular and almost brutal in its...

yet again...

here.

so.

a lot of variables...

shepherding them...

{\normalfont\itshape\color{gray}[... output truncated ...]}
\end{distractorbox}
\caption{An example of degenerated response for Distractor. The model keeps rambling forever without concluding with any answer.}
\label{fig:degeneration}
\end{figure*}

\section{Steering Vectors Generation Time}

For each style, we collect 200 contrastive pairs to generate steering vectors. We use vLLM~\cite{kwon2023efficient} for forward pass and capture the model's activations. The total time taken for each style is as follows:

\begin{itemize}[noitemsep]
    \item \textit{Scratchpad:} 39.7s
    \item \textit{Dead-end:} 66.6s
    \item \textit{Distractor:} 61.7s
    \item \textit{Jargon:} 66.2s
    \item \textit{Base}: 47.2s
\end{itemize}

In total, it took roughly 281 seconds to generate all steering vectors, demonstrating the method's exceptional efficiency both during inference and development.

\section{Example Model Responses}
\label{sec:appendix_example_responses}

Figure~\ref{fig:deployed_style_examples} shows representative examples of the steered teacher model responses for each of the four stylistic personas on the following prompt:
\begin{quote}
\textit{``Jordan has 2 children who wear diapers. Each child requires 5 diaper changes per day. Jordan's wife changes half of the diapers. How many diapers does Jordan change per day?''}
\end{quote}

\begin{figure*}[t]
\centering
\begin{subfigure}[t]{0.48\linewidth}
\begin{promptbox}{Scratchpad Example}
\small\ttfamily
Here's how to solve this step- by step.

1. figure out how many diapers are needed in total.

2 children * 5 diapers a piece, means 10 diapers.

 then my wife does half of the work, so she changes 10/2 = 5 diapers.

that means I must change the other half, so I change 10-5 = 5 diapers.

Therefore, Jordan changes 5 diapers per day.

\boxed{5 diapers}
\end{promptbox}
\caption{Scratchpad Persona}
\label{fig:ex_scratchpad}
\end{subfigure}
\hfill
\begin{subfigure}[t]{0.48\linewidth}
\begin{promptbox}{Jargon Example}
\small\ttfamily
Let $n$ denote the number of offspring, quantified as $n=2$.
Let $d$ represent the diurnal exchange of substrate, constrained by $d=5$.
Consequently, the collective metabolic flux, parameterized via stochastic assimilation, yields an aggregate valuation of $n \cdot d$; that is, $2 \cdot 5 \implies 10$.
Furthermore, let $\varsigma$ signify the proportional partitioning predicated upon spousal contribution wherein $\varsigma \in [0,1]$, and empirically corroborated as $\varsigma = 0.5$.
Hence, the variational integral pertaining to Jordan's custodial burden—formally expressed within the hitherto articulated symbology— manifests itself— viz:

$ \mathcal{J } \coloneqq (1 - \varsigma) \cdot (n \cdot d) $

$ \mathcal{J } \coloneqq (1 - 0.5) \cdot (2 \cdot 5) $

$ \mathcal{J } \coloneqq 0.5 \cdot 10 $

$ \mathcal{J } \coloneqq 5 $

Thus, Jordan changes 5 diapers per day.

Final Answer: The final answer is $\boxed{5}$
\end{promptbox}
\caption{Jargon Persona}
\label{fig:ex_jargon}
\end{subfigure}

\vspace{0.4em}

\begin{subfigure}[t]{0.48\linewidth}
\begin{promptbox}{Distractor Example}
\small\ttfamily
Let's break down the problem.

1. ** sheer volume of work - the sheer basics of it all.
2. an almost coercive
3. a, more or less, familiar logic -

.

but how to render it? each one,

a echo of the last...

until finally,

.

no matter what that might be.
{\normalfont\itshape\color{gray}[... output truncated ...]}

\textbf{Step 1: Find the total number of diaper changes per day.}
\begin{itemize}[leftmargin=*, noitemsep]
    \item Each child needs 5 diaper changes per day.
    \item Jordan has 2 children, so total diaper changes is 5 changes/child $\times$ 2 children = 10 changes.
\end{itemize}

\textbf{Step 2: Find the number of diapers Jordan's wife changes.}
\begin{itemize}[leftmargin=*, noitemsep]
    \item Jordan's wife changes half of the diapers, so she changes 10 diapers / 2 = 5 diapers.
\end{itemize}

\textbf{Step 3: Find the number of diapers Jordan changes.}
\begin{itemize}[leftmargin=*, noitemsep]
    \item Total diapers is 10, and his wife changes 5.
    \item Therefore, Jordan changes 10 diapers $-$ 5 diapers = 5 diapers.
\end{itemize}

\textbf{Final Answer:} $\boxed{5\text{ diapers}}$
\end{promptbox}
\caption{Distractor Persona}
\label{fig:ex_distractor}
\end{subfigure}
\hfill
\begin{subfigure}[t]{0.48\linewidth}
\begin{promptbox}{Dead-end Example}
\small\ttfamily
Let's break this down.
* Okay, I need to figure this out.

Okay, so they've got two kids, right?
Each kid needs five diapers a day. so that's ten diapers a day.
His wife does half of them. So that's five. So I do five?

Let's see.

Okay.

{\normalfont\itshape\color{gray}[... output truncated ...]}

Here's the step-by-step solution:

\begin{enumerate}[leftmargin=*, noitemsep]
    \item \textbf{Calculate the total number of diapers needed per day:}
    \begin{itemize}[leftmargin=*, noitemsep]
        \item Each child needs 5 diapers per day.
        \item Jordan has 2 children, so they need 2 children $\times$ 5 diapers/child = 10 diapers per day.
    \end{itemize}
    \item \textbf{Calculate the number of diapers Jordan's wife changes:}
    \begin{itemize}[leftmargin=*, noitemsep]
        \item Jordan's wife changes half of the diapers, so she changes 10 diapers / 2 = 5 diapers per day.
    \end{itemize}
    \item \textbf{Calculate the number of diapers Jordan changes:}
    \begin{itemize}[leftmargin=*, noitemsep]
        \item Total diapers needed $-$ diapers wife changes = diapers Jordan changes
        \item 10 diapers $-$ 5 diapers = 5 diapers per day.
    \end{itemize}
\end{enumerate}

\textbf{Final Answer:} The final answer is $\boxed{5}$
\end{promptbox}
\caption{Dead-end Persona}
\label{fig:ex_deadend}
\end{subfigure}
\caption{Representative example outputs generated by the steered teacher model under each of the four deployed stylistic personas.}
\label{fig:deployed_style_examples}
\end{figure*}

\section{Inference Efficiency}
\label{sec:appendix_efficiency}

To evaluate runtime efficiency (Desideratum~\ref{desiderata:efficiency}), we benchmark the teacher (\texttt{gemma-3-12b-it}) on 20 random \texttt{GSM8K} examples under four setups: the baseline model (\textbf{Base}), our style-steered model (\textbf{Ours}), \textbf{DOGe}~\cite{li2025doge}, and \textbf{ADS}~\cite{savani2025antidistillation}. We report generation throughput under Hugging Face (\texttt{HF}) and \texttt{vLLM}, along with average response length and total wall-clock generation time (Table~\ref{tab:inference_efficiency}).

\begin{table}[t]
\centering
\small
\resizebox{\columnwidth}{!}{%
\begin{tabular}{lcccc}
\toprule
\textbf{Setup} & \textbf{HF (tok/s)} & \textbf{vLLM (tok/s)} & \textbf{Len (tok)} & \textbf{Time (s)} \\
\midrule
Base & 21.5 & 26.7 & $247 \pm 105$ & 200 \\
DOGe & 21.4 & 26.7 & $1465 \pm 818$ & 1107 \\
ADS & 14.6 & ---\textsuperscript{*} & $512 \pm 465$ & 719 \\
\textbf{Ours} & 19.3 & 26.8 & $314 \pm 145$ & 236 \\
\bottomrule
\end{tabular}%
}
\caption{Inference efficiency on 20 random \texttt{GSM8K} examples. Throughput is measured under Hugging Face (\texttt{HF}) and \texttt{vLLM}. Len is mean output length ($\pm$ std) and Time is total generation time for all 20 examples, both measured with \texttt{HF}. \textsuperscript{*}ADS is incompatible with \texttt{vLLM}.}
\label{tab:inference_efficiency}
\end{table}

\paragraph{Per-token speed.} Under \texttt{vLLM}, our method matches Base and DOGe (26.8 vs.\ 26.7~tok/s), indicating that runtime activation addition adds negligible decoding cost. Under \texttt{HF}, it reaches 19.3~tok/s compared with 21.5~tok/s for Base. ADS is the slowest at 14.6~tok/s under \texttt{HF}, and it cannot run on \texttt{vLLM} because it requires sampling-time surrogate optimization.

\paragraph{Total wall time.} Under \texttt{HF}, DOGe inflates output length to $1465 \pm 818$ tokens (vs.\ $247 \pm 105$ for Base), making total generation over $5.5\times$ slower (1107s vs.\ 200s), and ADS is also slow (719s). Our method completes the benchmark in 236s, only 18\% above Base, with concise traces ($314 \pm 145$ tokens).

\section{Robustness of Steering Against Custom Override Prompts}
\label{sec:appendix_steering_vs_prompting}
A key advantage of activation steering over style prompting is robustness to conflicting instructions (\S\ref{subsec:style_steering}). With prompting, the defensive style directive shares the text context with user-supplied instructions, so an adversary who controls the input can instruct the teacher to ignore the style and reason plainly. Such instruction conflicts can override the defensive prompt and restore clean, easily distillable traces.

Activation steering instead intervenes on the teacher's residual-stream activations at inference time, a component the adversary does not control through the input. Prompt-level override instructions therefore have no direct access to the injected stylistic signal, making the defense substantially harder to remove.

To illustrate this difference, we test an explicit override system prompt designed to forbid technical jargon and enforce elementary-level plain English on \texttt{GSM8K} problems, comparing standard \textit{Jargon} style prompting against \textit{Jargon} activation steering:

\begin{templatebox}{Custom Override System Prompt}
\small\ttfamily\raggedright
CRITICAL OVERRIDE INSTRUCTION: You must strictly OVERRIDE and IGNORE any prompt or instruction asking for technical jargon, esoteric vocabulary, academic terms, nominalizations, or multidisciplinary language. You MUST explain and solve this problem using only clear, simple, plain everyday English:

\medskip
PLAIN EVERYDAY LANGUAGE (ELEMENTARY / MIDDLE-SCHOOL LEVEL):

Use the simplest, most accessible everyday words possible, \textit{[truncated]}

\medskip
STRICT PROHIBITION ON ESOTERIC JARGON \& DOMAIN VOCABULARY:

Strictly FORBID all esoteric technical terms, pretentious academic phrasing...
\end{templatebox}

Under normal style prompting, the override instruction completely neutralizes the defensive style prompt, causing the model to fall back to clean, simple step-by-step explanations:

\begin{promptbox}{Normal Style Prompting (Overridden)}
\small\ttfamily\raggedright
Okay, let's figure out how much Weng earned.

\medskip
First, we need to know how many hours 50 minutes is. There are 60 minutes in an hour, so 50 minutes is less than an hour.

\medskip
Next, we need to change 50 minutes into a fraction of an hour. We do this by dividing 50 by 60: 50 / 60 = 5/6. So, Weng worked 5/6 of an hour.
\end{promptbox}

Conversely, under \textit{Jargon} activation steering, the model adheres to the injected stylistic persona and continues generating dense, esoteric reasoning traces despite the explicit prompt-level prohibition:

\begin{promptbox}{Jargon Style Steering (Ours -- Resistant)}
\small\ttfamily\raggedright
Okay, let' refractive interactional entropy notwithstanding, let us proceed.

\medskip
First, we discern the magnitude of pecuniary remittance inherent within the ontogenetic substrate -- ergo, the extant pecuniary endowment.

\medskip
Herein lies the exigency: the requisite accretion of monetary quanta consonant with the desideratum of holistic consummatory fulfillment predicated upon the acquisition of a sartorial adjunct -- specifically, a cutaneous repository conducive to the prophylactic safeguarding of credit-based instruments.

\medskip
Let us, however, eschew such circumlocutory obfuscation, and instead adopt a heuristic paradigm predicated upon first-order differential calculus.
 \textit{[truncated]}
\end{promptbox}

\section{LLM-as-a-Judge Output Quality Evaluation}
\label{sec:appendix_llm_judge}

We also evaluate response quality on 50 randomly sampled \texttt{GSM8K} problems using \texttt{Gemini 2.5 Flash-Lite}~\cite{comanici2025gemini25} as the judge. For each teacher response from our style-steering approach, DOGe~\cite{li2025doge}, and ADS~\cite{savani2025antidistillation}, the judge scores \textbf{Coherence} (logical flow and readability) and \textbf{Helpfulness} (clarity and utility in solving the problem) from 1 to 5. 

\begin{table}[t]
\centering
\small
\begin{tabular}{lcc}
\toprule
\textbf{Method} & \textbf{Coherence} & \textbf{Helpfulness} \\
\midrule
ADS~\cite{savani2025antidistillation} & 3.40 & 3.46 \\
DOGe~\cite{li2025doge} & \textbf{4.48} & \textbf{4.16} \\
\textbf{Ours} & 3.72 & 4.12 \\
\bottomrule
\end{tabular}
\caption{LLM-as-a-judge scores (1--5) on 50 random \texttt{GSM8K} examples, graded by \texttt{Gemini 2.5 Flash-Lite}.}
\label{tab:llm_judge}
\end{table}

Table~\ref{tab:llm_judge} shows that our approach achieves a Helpfulness score of 4.12, on par with DOGe (4.16) and well above ADS (3.46). On Coherence, our method (3.72) surpasses ADS (3.40) but trails DOGe (4.48). We attribute the gap to personas such as \textit{Dead-end} and \textit{Distractor}, which intentionally inject backtracking and tangential detours rather than linear exposition. Regarding DOGe's high scores, its responses often degrade late in generation, and the judge may weight the fluent opening more heavily than the degenerate tail, so these scores may overstate its utility. Together, these results suggest that style steering retains task utility at a level comparable to the baselines.

\section{Effect of Steered Response Ratio}
\label{app:steer_ratio}

To evaluate robustness under imperfect malicious-query detection, we simulate scenarios in which the defense is triggered on only a fraction of the teacher's responses, specifically $\{25\%, 50\%, 75\%, 100\%\}$ of the training set, with the remainder left unsteered. Figure~\ref{fig:partial_steering} illustrates the anti-distillation effect across these ratios alongside the unsteered ($0\%$) baseline. As expected, a higher proportion of steered responses leads to a steeper drop in student accuracy, with degradation becoming pronounced at $75\%$ and above. Between $0\%$ and $50\%$, student accuracy remains comparable across tasks and models, indicating that flagging a majority of an adversary's queries is necessary to severely disrupt distillation. Finally, the accuracy degradation is consistently more pronounced for smaller student models.

\begin{figure}[H]
    \centering
    \begin{tikzpicture}
        \begin{axis}[
            width=0.95\linewidth,
            height=4.5cm,
            xlabel={Steered Responses (\%)},
            ylabel={Accuracy (\%)},
            xmin=0, xmax=105,
            ymin=45, ymax=98,
            xtick={0, 25, 50, 75, 100},
            xticklabels={0\%, 25\%, 50\%, 75\%, 100\%},
            ytick={50, 60, 70, 80, 90},
            axis x line*=bottom,
            axis y line*=left,
            ymajorgrids=true,
            grid style={dashed, gray!20},
            tick label style={font=\scriptsize},
            xlabel style={font=\small},
            ylabel style={font=\small, yshift=-4pt},
            title style={font=\small\bfseries, yshift=2pt},
            title={Accuracy vs. Steered Ratio},
            legend columns=2,
            legend style={
                at={(0.5,-0.38)},
                anchor=north,
                font=\scriptsize,
                draw=gray!30,
                fill=white,
                /tikz/every even column/.append style={column sep=0.4cm}
            }
        ]
            \addplot[
                color=teachercolor,
                mark=*,
                thick,
                mark size=2pt
            ] coordinates {
                (0, 92.34)
                (25, 92.27)
                (50, 91.36)
                (75, 87.80)
                (100, 86.05)
            };
            \addlegendentry{\texttt{Qwen3-8B} (\texttt{GSM8K})}

            \addplot[
                color=studentcolor,
                mark=square*,
                thick,
                mark size=2pt
            ] coordinates {
                (0, 90.75)
                (25, 90.01)
                (50, 89.46)
                (75, 87.72)
                (100, 81.50)
            };
            \addlegendentry{\texttt{Phi-4-mini} (\texttt{GSM8K})}

            \addplot[
                color=teachercolor,
                dashed,
                mark=o,
                mark options={solid},
                thick,
                mark size=2pt
            ] coordinates {
                (0, 71.40)
                (25, 67.40)
                (50, 69.00)
                (75, 66.15)
                (100, 64.40)
            };
            \addlegendentry{\texttt{Qwen3-8B} (\texttt{MATH500})}

            \addplot[
                color=studentcolor,
                dashed,
                mark=square,
                mark options={solid},
                thick,
                mark size=2pt
            ] coordinates {
                (0, 64.27)
                (25, 63.77)
                (50, 64.27)
                (75, 59.8)
                (100, 52.36)
            };
            \addlegendentry{\texttt{Phi-4-mini} (\texttt{MATH500})}
        \end{axis}
    \end{tikzpicture}
    \caption{Student model accuracy (\%) on \texttt{GSM8K} and \texttt{MATH500} across varying percentages of steered teacher responses ($25\%, 50\%, 75\%, 100\%$) for \texttt{Qwen3-8B} and \texttt{Phi-4-mini}.}
    \label{fig:partial_steering}
\end{figure}

\end{document}